%% file: _main.tex
\def\cvprPaperID{12789}
\def\confName{CVPR}
\def\confYear{2023}

\def\paperTitle{Regularize implicit neural representation by itself}

\def\authorBlock{
	Zhemin Li$^1$
	\qquad
	Hongxia Wang\thanks{This work was supported by the National Key Research ,Development Program (2020YFA0713504), the National Natural Science Foundation of China (61977065) and the Macao Science and Technology Development Fund (061/2020/A2).} $^1$ \qquad
	Deyu Meng$^{2,3}$
	\\
	$^1$  National University of Defense Technology \quad
	$^2$  Xi'an Jiaotong University \\ $^3$ Macau University of Science and Technology \\
	{\tt\small lizhemin@nudt.edu.cn,wanghongxia@nudt.edu.cn,dymeng@mail.xjtu.edu.cn}
}

\newif\ifreview 
\newif\ifarxiv \newcommand{\arxiv}{\arxivtrue}
\newif\ifcamera 
\newif\ifrebuttal 

\arxiv 

\pdfoutput=1
\documentclass[10pt,twocolumn,letterpaper]{article}
\input{cvpr_header.tex}
\input{_math.tex}

\newcommand{\Min}{\textrm{minimize}}
\usepackage{multirow}
\usepackage{lipsum}
\begin{document}
\title{\paperTitle}
\author{\authorBlock}
\maketitle
\newcommand\blfootnote[1]{%
\begingroup
\renewcommand\thefootnote{}\footnote{#1}%
\addtocounter{footnote}{-1}%
\endgroup
}

\begin{abstract}
	This paper proposes a regularizer called Implicit Neural Representation Regularizer (INRR) to improve the generalization ability of the Implicit Neural Representation (INR). The INR is a fully connected network that can represent signals with details not restricted by grid resolution. However, its generalization ability could be improved, especially with non-uniformly sampled data. The proposed INRR is based on learned Dirichlet Energy (DE) that measures similarities between rows/columns of the matrix. The smoothness of the Laplacian matrix is further integrated by parameterizing DE with a tiny INR. INRR improves the generalization of INR in signal representation by perfectly integrating the signal's self-similarity with the smoothness of the Laplacian matrix. Through well-designed numerical experiments, the paper also reveals a series of properties derived from INRR, including momentum methods like convergence trajectory and multi-scale similarity. Moreover, the proposed method could improve the performance of other signal representation methods.
\end{abstract}

\section{Introduction}
\label{sec:intro}
\begin{figure*}[htb]
	\centering
	\includegraphics[width=0.7\linewidth]{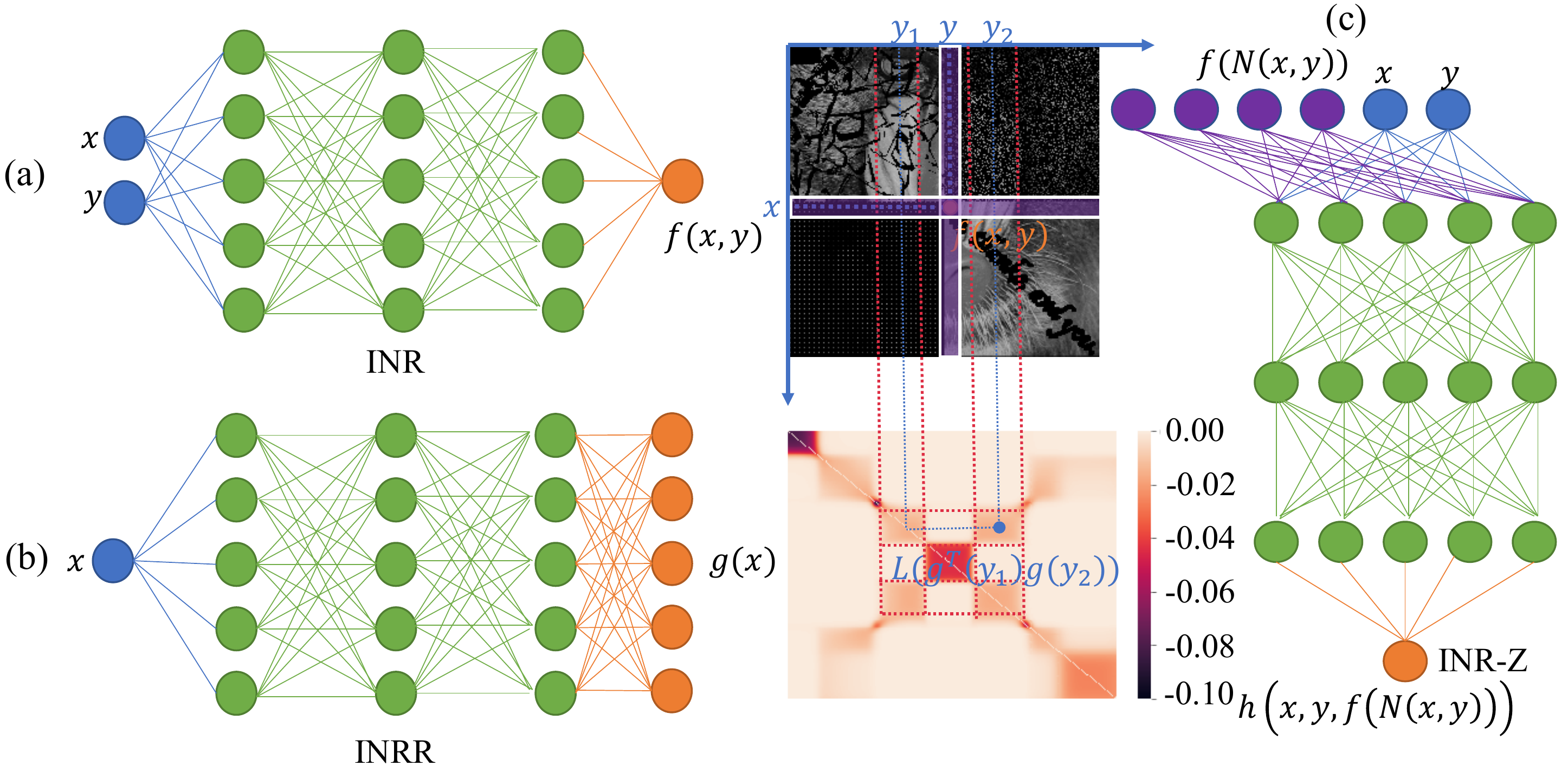}
	\caption{Overview of proposed improve scheme for INR. (a) INR is a fully connected neural network which maps from coordinate to pixel value. (b) INRR is a regularization term represented by an INR which can capture the self-similarity. (c) INR-Z improve the performance of INR by combining the neighbor pixels with coordinate together as the input of another INR.}
	\label{fig:structure}
\end{figure*}

\begin{figure}[htb]
	\begin{center}
		\begin{tabular}{ccc}
			\hspace{-0.3cm}\includegraphics[width=0.31\columnwidth]{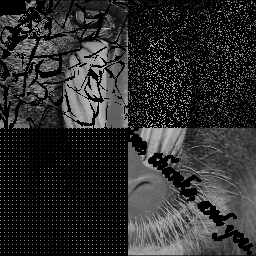}&
			\hspace{-0.35cm}\includegraphics[width=0.31\columnwidth]{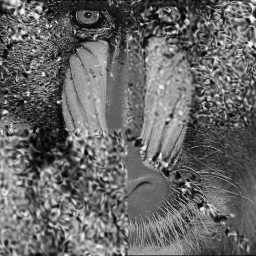}&
			\hspace{-0.35cm}\includegraphics[width=0.31\columnwidth]{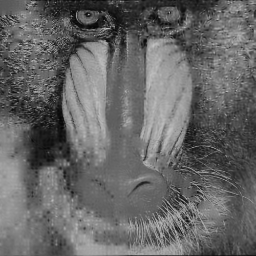}\\
			{\footnotesize (a) Sampling} &\hspace{-0.3cm} {\footnotesize (b) INR (18.1 dB)} &\hspace{-0.3cm}  {\footnotesize (c) INRR (23.3 dB)} 
		\end{tabular}
	\end{center}\vspace{-0.3cm}
	\caption{Image fitting results. All the methods are based on the SIREN to fit an $256\times 256$ Baboon with the sampling data in (a). (b) trained with a vanilla SIREN while (c) trained with proposed INRR.}\label{fig:Whenbad}
\end{figure}

INR uses a fully connected network (FCN) $\phi_{\boldsymbol{\theta}}(\mathbf{x}):\mathbb{R}^d\mapsto\mathbb{R}^o$ to approximate the explicit solution of an implicit function $F\left(\mathbf{x},\phi_{\boldsymbol{\theta}},\nabla_\mathbf{x} \phi_{\boldsymbol{\theta}},\nabla_\mathbf{x}^2 \phi_{\boldsymbol{\theta}},\ldots\right)=0$. For an example, we can represent a gray-scale image $\mathbf{X}\in\mathbb{R}^{m\times n}$ with an INR $\phi_{\boldsymbol{\theta}}(\mathbf{x}):\mathbb{R}^2\mapsto\mathbb{R}$ which satisfied $\phi_{\boldsymbol{\theta}}(\frac{i}{m},\frac{j}{n})=\mathbf{X}_{ij},
i\in \left\{1,\ldots,m\right\}, j\in\left\{1,\ldots,n\right\}$. Compared with traditional grid representation $\mathbf{X}$, INR's representation ability to details is not restricted by grid resolution $m,n$ as INR can predict the pixel value at any location $(x,y)\in\mathbb{R}^2$ even not equals to $(\frac{i}{m},\frac{j}{n})$.

Besides the representation ability of INR, generalization ability is critical for a neural network. We explore the empirical generalization ability via a $256\times 256$ gray-scale non-uniformly sampled image inpainting task as \Figref{fig:Whenbad}(a) shows. Although INR fits training data perfectly in \Figref{fig:Whenbad}(b), its prediction outside training data is unreasonable. Theoretical analysis of INR illustrates that a hyper-parameter controls the smoothness degree of $\phi_{\boldsymbol{\theta}}(\mathbf{x})$. Moreover, the experiments show that the best hyper-parameter varies with the missing rate (the percentage of unsampled pixels) as \Figref{fig:inrk} shows. Adjusting this hyper-parameter cannot make the non-uniformly missing case perform best, as different locations might have different missing rates.

A carefully designed regularizer is proposed to improve the generalization ability of INR. It is based on Adaptive and Implicit Regularization (AIR) which is a learned Dirichlet Energy (DE)\cite{ZheminLi2022AdaptiveAI} that measures similarities or correlations between rows/columns of $\mathbf{X}$. The smoothness of the Laplacian matrix is further integrated by parameterizing DE with a tiny INR. The structure of the proposed implicit neural representation regularizer (INRR) is shown in \Figref{fig:structure}(b). Because a smooth Laplacian matrix represents non-local prior and large-scale local prior in vision data, INRR can improve the generalization of INR in image representation. Numerous numerical experiments show that INRR outperforms various classical regularizers, including total variation (TV), $L_2$ energy, and so on. As a regularizer both in a new form and with new meaning, INRR can be combined with other signal representation methods, such as deep matrix factorization (DMF) \cite{SanjeevArora2019ImplicitRI}.

To summarize, the contributions of our work include the following:
\begin{itemize}
	\setlength{\itemsep}{-1pt}
	\item Neural Tangent Kernel (NTK) \cite{SanjeevArora2019ImplicitRI} theoretically analyzes the generalization ability of INR and why INR performs poorly with nonuniform sampling is given.
	\item A tiny INR parameterized regularizer named INRR is proposed based on DE, which perfectly integrates the image's self-similarity with the smoothness of the Laplacian matrix.
	\item A series of properties derived from INRR, including momentum methods, multi-scale similarity, and generalization ability, are revealed by well-designed numerical experiments. 
\end{itemize}

\section{Related Work}
\label{sec:related}
\textbf{Implicit neural representation}. 
Recently, INR has shown outstanding potential in representing vision data, including font, images, and videos \cite{PradyumnaReddy2021AMN,VincentSitzmann2020ImplicitNR}. It has been applied in novel view synthesis \cite{BenMildenhall2020NeRFRS,RicardoMartinBrualla2020NeRFIT,KrishnaWadhwani2022SqueezeNeRFFF,AjayJain2021PuttingNO}, signal compression \cite{YunfanZhang2022ImplicitNV,YannickStrmpler2022ImplicitNR,EmilienDupont2021COINCW,EmilienDupont2022COINNC,FrancescaPistilli2022SIGNALCV}, and classification\cite{EmilienDupont2022FromDT,IshitMehta2021ModulatedPA}. 

In these latter years, a series of works have systematically studied and advanced the representation capabilities of INR. Tancik et al. discuss why an INR with ReLU activation function can not represent the high-frequency components well and introduce a Fourier feature encode that significantly improves the representation ability of INR \cite{MatthewTancik2020FourierFL}. Furthermore, Stizmann et al. replace ReLU with a sinuous activation function and propose a specific initialization scheme. The corresponding network is named sinusoidal representation network (SIREN) \cite{VincentSitzmann2020ImplicitNR}.  
Then Fathony et al. propose filter neural networks with the Fourier and Gabor as basis activation \cite{Fathony2021MultiplicativeFN}. Furthermore, Band-limited Coordinate Networks (BACON) introduces the ability of multiscale INR representation \cite{DavidBLindell2022BACONBC}. Apart from fitting the training set, the generalization ability of INR is more critical in many applications.

\textbf{Regularization.} 
Improving the generalization of NN with regularization techniques such as $L_1$-norm, $L_2$-norm, and the Dropout technique has a long history \cite{Srivastava2014DropoutAS}. These regularizations take the images or other signals as input. Recently, there has been a class of NN that use a whole NN to represent a signal, such as Deep Image Prior (DIP), Deep Matrix Factorization (DMF), and INR \cite{Ulyanov2018DeepIP,SanjeevArora2019ImplicitRI,VincentSitzmann2020ImplicitNR}. In this case, the classical signal regularization technique can be applied to the signal represented NN \cite{Liu2019ImageRU,GaryMataev2019DeepREDDI,ZheminLi2022AdaptiveAI,Li2022IET}. Significantly, the learnable regularizer is better than those not learned \cite{GaryMataev2019DeepREDDI,ZheminLi2022AdaptiveAI}. To our knowledge, no effort has been made to regularize INR using a learnable regularizer based on the characteristics of INR's data representation.

\section{Theoretical analysis of INR}
\label{sec:theory}
\begin{figure*}[htb]
	\centering
	\includegraphics[width=0.9\linewidth]{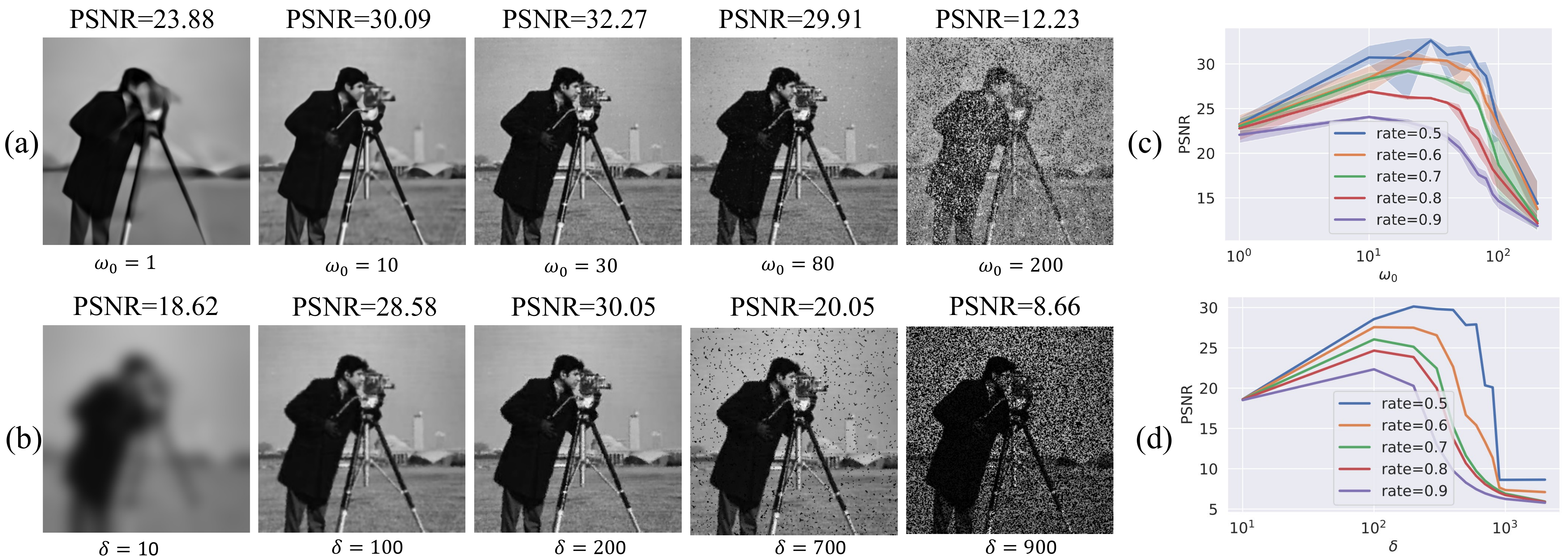}
	\caption{Fitting a $256\times 256$ Cameraman which random missing $50\%$ pixels with (a) SIREN and (b) NTK, respectively. $\omega_0$ and $\delta$ are the hyper-parameters of models. (c,d) shows the PSNR value change with $\omega_0,\delta$ at different random missing rate, respectively.}
	\label{fig:inrk}
\end{figure*}

As \Figref{fig:Whenbad}(b) shows, INR's generalization ability is not as well as its representation ability. We analyze INR theoretically with a proxy model NTK to answer when and why INR generalizes badly.

\textbf{Implicit neural representation.}
INR uses a FCN $\phi_{\boldsymbol{\theta}}(\mathbf{x}):\mathbb{R}^d\mapsto\mathbb{R}^o$ to approximate the explicit representation of an implicit function $F\left(\mathbf{x},\phi_{\boldsymbol{\theta}},\nabla_\mathbf{x} \phi_{\boldsymbol{\theta}},\nabla_\mathbf{x}^2 \phi_{\boldsymbol{\theta}},\ldots\right)=0$, where the FCN has $L$ hidden layers defined as follows,
\begin{equation}
	\label{eq:inr}
	\begin{aligned}
		\mathbf{z}^{(\ell)}=\mathbf{W}^{(\ell)} \mathbf{x}^{(\ell-1)}+\mathbf{b}^{(\ell)}&, \quad \mathbf{x}^{(\ell)}=\sigma\left(\mathbf{z}^{(\ell)}\right),\\
		\phi_{\boldsymbol{\theta}}(\mathbf{x})=\mathbf{z}^{(L+1)}&,\quad \ell=1,2,\ldots,L+1,
	\end{aligned}
\end{equation}
with $\sigma(\cdot)$ an element-wise activation function and $\mathbf{x}^{(0)}=\mathbf{x}$, $\boldsymbol{\theta}=\left\{\mathbf{W}^{(\ell)},\mathbf{b}^{(\ell)}\mid \ell=1,2,\ldots,L+1\right\}\sim \mathcal{D}$ at initialization, $\mathbf{W}^{(\ell)}\in\mathbb{R}^{n_\ell\times n_{\ell-1}}$, and $n_\ell$ is the width of $\ell$-th layer with $n_0=d$, $n_{L+1}=o$.

For simplicity, we focus on a special case of INR with $F(\boldsymbol{\theta},\mathbf{x},z)=\left\|\phi_{\boldsymbol{\theta}}(\mathbf{x})-z(\mathbf{x})\right\|_2^2=0$ and $\phi_{\boldsymbol{\theta}}(\mathbf{x}):\mathbb{R}^2\mapsto \mathbb{R}$, where $z(\mathbf{x}):\mathbb{R}^2\mapsto \mathbb{R}$ is the gray-scale image we want to represent. All the results in this paper can be easily extended to a higher dimension. The vanilla INR is formulated as 
\begin{equation}
	\boldsymbol{\theta}^*=\arg\min_{\boldsymbol{\theta}}\left\{\mathcal{L}(\boldsymbol{\theta},\mathcal{X},\mathcal{Z})=\sum_{(\mathbf{x}_i, z_i)\in\mathcal{X}\times \mathcal{Z}}F\left(\boldsymbol{\theta},\mathbf{x}_i,z_i\right)\right\}, 
\end{equation}
where $\mathcal{X}\times \mathcal{Z}=\left\{(\mathbf{x}_i, z_i)\right\}_{i=1}^N$ is the training set, and $\mathbf{x}_i=(x_i,y_i)$ is the coordinate. The training set is sampled from the grid of matrix $\mathbf{X}\in\mathbb{R}^{m\times n}$. For example, we can use $(\frac{i}{m},\frac{j}{n})$ as input and $\mathbf{X}_{ij}$ as the corresponding output of INR. After training, $z(\mathbf{x})$ is predicted by $\phi_{\boldsymbol{\theta}}(\mathbf{x})$ at any location $\mathbf{x}=(x,y)$ even when $\mathbf{x} \not\in \mathcal{G}=\left\{(\frac{i}{m},\frac{j}{n})\mid i\in\left\{1,\ldots,m\right\},j\in\left\{1,\ldots,n\right\}\right\}$.

\textbf{Kernel regression approximate neural networks.}
Jacot et al. show that with infinity width of the layers in $\phi_{\boldsymbol{\theta}}$ and small learning rate, the function $\phi_{\boldsymbol{\theta}}$ converges to the kernel regression
$$
\phi_{\text{NTK}}(\mathbf{x})=\sum_{i=1}^N \left(\mathbf{K}^{-1}\mathbf{z}\right)_i k_{\text{NTK}}\left(\mathbf{x}_i,\mathbf{x}\right),
$$
where $\mathbf{K}$ is an $N\times N$ kernel matrix dubbed neural tangent kernel (NTK) \cite{ArthurJacot2018NeuralTK,EugeneGolikov2022NeuralTK,GregYang2020TensorPI} with entries defined as  
$$
\mathbf{K}_{ij}=k_{\text{NTK}}(\mathbf{x}_i,\mathbf{x}_j)= \mathbb{E}_{\boldsymbol{\theta}\sim \mathcal{D}} \left<\frac{\partial \phi_{\boldsymbol{\theta}}(\mathbf{x}_i)}{\partial \boldsymbol{\theta}},\frac{\partial \phi_{\boldsymbol{\theta}}(\mathbf{x}_j)}{\partial \boldsymbol{\theta}}\right>.
$$

In this paper, we consider INR $\phi':\mathbb{R}^{2D}\mapsto \mathbb{R}$ with a feature map $\gamma(\mathbf{x})=\frac{1}{\sqrt{D}}[\cos \mathbf{Bx}^\top,\sin \mathbf{Bx}^\top]^\top:\mathbb{R}^d\mapsto \mathbb{R}^{2D}$ as its input, where $\mathbf{x}\in \mathbb{R}^{d}$, $\mathbf{B}\in \mathbb{R}^{D\times d}$, and $\mathbf{B}_{ij}\sim \mathcal{N}(0,\delta)$. Then $\phi_{\text{NTK}}'(\gamma(\mathbf{x}))$ is shift-invariant thus more suitable for image representation.

Now we analyze how INR predicts the data outside of the training set. Theorem \ref{thm:infd} illustrates that the smoothness of represented signal is controlled by the hyper-parameter $\delta$ globally. Especially when $\delta$ tends to infinity, the prediction of $\phi_{\text{NTK}}'(\cdot)$ outside the training set all tends to the same weighted average of the training set according to Corollary \ref{cor}. 

\begin{thm}
	\label{thm:infd}
	Given a FCN $\phi_{\boldsymbol{\theta}}'(\cdot):\mathbb{R}^{2D}\mapsto \mathbb{R}$ with $\mathbf{x}\in\mathbb{R}^d$, $\gamma(\mathbf{x})\in\mathbb{R}^{2D}$, and the feature map $\gamma(\mathbf{x})=\frac{1}{\sqrt{D}}[\cos \mathbf{Bx}^\top,\sin \mathbf{Bx}^\top]^\top$ with $\mathbf{B}\in \mathbb{R}^{D\times d}$ and $\mathbf{B}_{ij}\sim \mathcal{N}(0,\delta)$. Denote the corresponding composed NTK as $k_D(\mathbf{x}_i,\mathbf{x}_j)=h_{\text{NTK}}\left(\frac{1}{D}\mathbf{1}_D^\top\cos\left(\mathbf{B}(\mathbf{x}_i-\mathbf{x}_j)\right)\right)$, then we have
	$$
	\lim_{D\rightarrow \infty} k_D(\mathbf{x}_j,\mathbf{x}_j)=h_{\text{NTK}}(e^{-\delta^2 \left\|\mathbf{x}_i-\mathbf{x}_j\right\|^2}).
	$$
\end{thm}

\begin{cor}
	\label{cor}
	Assume the $h_{\text{NTK}}$ in Theorem \ref{thm:infd} satisfies $h_{\text{NTK}}(1)\neq h_{\text{NTK}}(0)$ and $h_{\text{NTK}}(1)\neq 0$, then
	$$
	\lim_{\delta\rightarrow \infty} \phi_{\text{NTK}}'(\gamma(\mathbf{x}))=\left\{\begin{array}{cc}
	z_l,\qquad \quad \mathbf{x}=\mathbf{x}_l\in\left\{\mathbf{x}_i\right\}_{i=1}^N,\\
	\frac{h(0)\mathbf{1}_N^\top\mathbf{z}}{(N-1)h(0)+h(1)},\mathbf{x}\not\in\left\{\mathbf{x}_i\right\}_{i=1}^N.\\
	\end{array}
	\right.
	$$
\end{cor}

\textbf{INR needs to be regularized.}
We validate Corollary \ref{cor} by exploring the performance of $\phi_{\text{NTK}}'(\gamma(\mathbf{x}))$ in image inpainting task with different missing rates. As \Figref{fig:inrk}(b) shows, when $\delta=900$, $\phi_{\text{NTK}}'(\gamma(\mathbf{x}))$ at the location of outside of sampled data has the same value. Furthermore, \Figref{fig:inrk}(a) shows that the latest SIREN \cite{VincentSitzmann2020ImplicitNR}, which represents signals without a feature map of input, is also controlled by the hyper-parameter $\omega_0$ in the first layer as $\sin\left(\omega_0 \mathbf{W}\mathbf{x}+\mathbf{b}\right)$.

Based on the numerical result, the optimal $\omega_0$ or $\delta$ is required so that INR generalizes the best. However, finding an optimal $\omega_0$ or $\delta$ with non-uniformly sampled training data is impossible. \Figref{fig:inrk}(c,d) illustrates that the optimal $\omega_0$ or $\delta$ varies considerably according to the missing rate. It decreases with the increase of missing rate, which is consistent with the theoretical results that the sparser sampling needs a smoother fitting. As to the case with nonuniform missing, note that different locations might have different missing rate; it is tough to make INR performs well by choosing an optimal hyper-parameter.  

Furthermore, the results above all based on the loss function $\mathcal{L}=\sum_{(\mathbf{x}_i,z_i)\in \mathcal{X}\times \mathcal{Z}} \left\|\phi_{\boldsymbol{\theta}}(\mathbf{x}_i)-z_i\right\|_2^2$ which is a fidelity term measured on the training data. Enforcing additional constraints on the predicted data is profitable to improve the generalization ability of INR. In the next section, we add constraints by a newly proposed regularizer named INRR.

\section{Methods to regularize INR}
\label{sec:method}
This section presents a regularized model $\mathcal{L}(\boldsymbol{\theta}, \mathcal{X}, \mathcal{Z})+\lambda \mathcal{R}(\boldsymbol{\theta}, \mathcal{X}, \mathcal{Z})$, where $\lambda$ is a parameter that balances the loss of training data and the regularizer $\mathcal{R}$. 

Now consider the priors of images on a larger scale. Since the vanilla INR's loss function is pixel-by-pixel, it ignores the structural features of images. Specifically, these features include the relationship between rows, columns, or blocks. Low rank is a well-known prior that describes the correlation between rows and columns. However, a low-rank matrix cannot express the details of a signal well because these details are located in the subspaces corresponding to the small singular value of the image. 

So we turn to self-similarity, which is quite common in large and fine scales of an image. As a simple example, smooth $\mathbf{X}$ implies local similarity between adjacent rows and columns of $\mathbf{X}$. Furthermore, the non-local self-similarity of an image, which refers to the similarity between non-adjacent rows, columns, or blocks, is also very universal and valuable. In this paper, we choose Dirichlet Energy (DE) to describe images' local and non-local self-similarity. Our method is not restricted to DE. 

\subsection{Dirichlet Energy}
Given a matrix $\mathbf{X}\in\mathbb{R}^{m\times n}$, DE is formulated as follows
$$
\mathcal{R}_{\text{DE}}=tr\left(\mathbf{X}^\top \mathbf{L}\mathbf{X}\right)=\sum_{1\leq i,j\leq m} \mathbf{A}_{ij}\left\|\mathbf{X}_{i,:}-\mathbf{X}_{j,:}\right\|^2,
$$
where $\mathbf{A}\in\mathbb{R}^{m\times m}$ is a weighted adjacency matrix along rows of $\mathbf{X}$, and $\mathbf{L}=\mathbf{D}-\mathbf{A}$ with $\mathbf{D}_{ii}=\sum_{j=1}^m \mathbf{A}_{ij}$ and $\mathbf{D}_{ij}=0$ if $i\neq j$. As $\mathbf{A}_{ij}$ measures the similarity of rows $\mathbf{X}_{i,:}$ and $\mathbf{X}_{j,:}$,  DE is a non-local self-similarity measure of $\mathbf{X}$. 

However, there are two main issues in using DE: (a) $\mathbf{L}$ or $\mathbf{A}$ is unknown under the incomplete sampling of $\mathbf{X}$; (b) DE only encodes the similarity between two rows, other large-scale similarities such as block similarity cannot be captured. To solve these problems, we parameterize $\mathbf{L}$ with another tiny INR and learn it during training $\phi_{\boldsymbol{\theta}}(\mathbf{x})$. 

\subsection{INRR}
\label{subsec:inrr}
Learning $\mathbf{L}$ during training is naive thinking when $\mathbf{L}$ is unknown. Nevertheless, we need to sufficiently extract the properties of $\mathbf{L}$ to make it meaningful and practical. There are two mathematical properties that $\mathbf{L}$ needs to satisfy: (a) positive semi-definite, (b) the sum of each row equals zero. Specially, we find the $\mathbf{L}$ of natural images has some extra priors. The natural images are usually piecewise smooth, so $\mathbf{L}$, which measures the similarity of the rows of $\mathbf{X}$, should also be nearly smooth.

Therefore, we propose an implicit neural representation regularization (INRR) which is expressed as follows:
$$
\left\{\begin{array}{c}
\begin{aligned}
\mathcal{R}(\boldsymbol{\theta})&=tr\left(\left[\mathcal{T}(\mathbf{X})\right]^\top \mathbf{L}(\boldsymbol{\theta})\mathcal{T}(\mathbf{X})\right) \\
\mathbf{L}(\boldsymbol{\theta})&=\mathbf{A}(\boldsymbol{\theta})\cdot \mathbf{1}_{m'\times m'}\odot \mathbf{I}_{m'}-\mathbf{A}(\boldsymbol{\theta})\\
\mathbf{A}(\boldsymbol{\theta}) &= \frac{\text{exp}\left(g^\top(\boldsymbol{\theta};\mathbf{u})g(\boldsymbol{\theta};\mathbf{u})\right)}{\mathbf{1}_{m'}^\top \text{exp}\left(g^\top(\boldsymbol{\theta};\mathbf{u})g(\boldsymbol{\theta};\mathbf{u})\right)\mathbf{1}_{m'}}
\end{aligned}
\end{array}
\right.,
$$
where $\mathcal{T}(\cdot):\mathbb{R}^{m\times n}\mapsto \mathbb{R}^{m'\times n'}$ aims to capture self-similarity in $\mathbf{X}$, $\mathbf{L}(\boldsymbol{\theta})$ measures the similarity between rows of $\mathcal{T}(\mathbf{X})$. $g(\boldsymbol{\theta};\cdot):\mathbb{R}\mapsto \mathbb{R}^r$ is a tiny INR, $g(\boldsymbol{\theta};\mathbf{u})\in\mathbb{R}^{r\times m'}$, $g(\boldsymbol{\theta};\mathbf{u})_i=g(\boldsymbol{\theta};\mathbf{u}_i),i=1,2,\ldots,m'$. And $\mathbf{u}_i$ is coordinate of sampled matrix $\mathcal{T}(\mathbf{X})$ with $\mathcal{T}(\mathbf{X})_{ij}=\phi_{\boldsymbol{\theta}}\left(\mathbf{u}_i,\mathbf{v}_j\right)$ and $\mathbf{u}=\left[\frac{1}{m'},\frac{2}{m'},\ldots,\frac{m'}{m'}\right]^\top$, $\mathbf{v}=\left[\frac{1}{n'},\frac{2}{n'},\ldots,\frac{n'}{n'}\right]^\top$. It is not difficult to verify that the parameterized Laplacian matrix keeps properties (a) and (b). Furthermore, $g(\boldsymbol{\theta};\cdot)$ introduces the smoothness of $\mathbf{L}$ implicitly, and $r$ restricts the rank of $\mathbf{L}$.

Take the relations between columns into account simultaneously. The whole regularized model is formulated as
$$
\Min_{\boldsymbol{\theta},\boldsymbol{\theta}_r,\boldsymbol{\theta}_c} \left\{\mathcal{L}(\boldsymbol{\theta},\mathcal{X},\mathcal{Z})+\lambda_r \mathcal{R}(\boldsymbol{\theta}_r)+\lambda_c \mathcal{R}(\boldsymbol{\theta}_c)\right\},
$$
where $\mathcal{R}(\boldsymbol{\theta}_r)$ and $\mathcal{R}(\boldsymbol{\theta}_c)$ are row and column regularizers respectively. $\mathcal{T}(\mathbf{X})=\mathbf{X}$ in $\mathcal{R}(\boldsymbol{\theta}_r)$, and $\mathcal{T}(\mathbf{X})=\mathbf{X}^\top$ in $\mathcal{R}(\boldsymbol{\theta}_c)$. $\lambda_r,\lambda_c$ are used to balance the fidelity and regularization terms.

As the self-similarity which is represented by $g(\boldsymbol{\theta}_r;\cdot)$ or $g(\boldsymbol{\theta}_c;\cdot)$ are much simpler than the image, so the parameter number of $g(\boldsymbol{\theta}_r;\cdot)$ or $g(\boldsymbol{\theta}_c;\cdot)$ are much lesser than the one of $\phi_{\boldsymbol{\theta}}(\cdot)$, which called tiny INR.

\section{Experiments}
\label{sec:exp}
\subsection{Experimental setting}
\begin{table*}[ht]
	\tabcolsep=0.15cm
	\caption{PSNR (dB) of recovered images by INR based models with different missing patters include random missing, patch missing, and textural missing. Four images are tested including Baboon, Barbara, Man and Boats.}
	\begin{tabular}{lcllllllcllllll}
		\hline
		& \multicolumn{1}{l}{} & INR & INR-Z & TV & $L_2$ & AIR & \textbf{INRR} & \multicolumn{1}{l}{} & INR & INR-Z & TV & $L_2$ & AIR & \textbf{INRR} \\ \hline
		Random & \multirow{3}{*}{Baboon} & 21.0 & 21.5 & 23.8 & 23.1 & 24.2 & \textbf{24.7} & \multirow{3}{*}{Barbara} & 28.6 & 28.8 & 30.2 & 30.5 & 30.3 & \textbf{30.7} \\
		Patch &  & 25.3 & 27.3 & 24.5 & 26.3 & 31.6 & \textbf{33.8} &  & 26.1 & 27.1 & 28.8 & 28.5 & 28.9 & \textbf{29.5} \\
		Textural &  & 20.8 & 23.5 & 21.4 & 23.9 & 27.2 & \textbf{28.3} &  & 26.5 & 27.9 & 26.5 & 27.9 & 29.0 & \textbf{29.4} \\ \hline
		Random & \multirow{3}{*}{Man} & 23.5 & 22.8 & 25.6 & 25.6 & 25.7 & \textbf{25.9} & \multirow{3}{*}{Boats} & 28.1 & 27.6 & 29.5 & 29.7 & 29.4 & \textbf{29.9} \\
		Patch &  & 25.6 & 26.5 & 22.7 & 26.9 & 31.3 & \textbf{32.1} &  & 27.6 & 28.5 & 27.8 & 29.5 & 33.4 & \textbf{34.4} \\
		Textural &  & 22.6 & 22.6 & 24.2 & 24.9 & 25.3 & \textbf{26.4} &  & 24.1 & 27.0 & 25.2 & 28.1 & 27.9 & \textbf{28.9} \\ \hline
	\end{tabular}
	\label{tab:diffreg}
\end{table*}
\begin{figure*}[htb]
	\centering
	\includegraphics[width=0.8\linewidth]{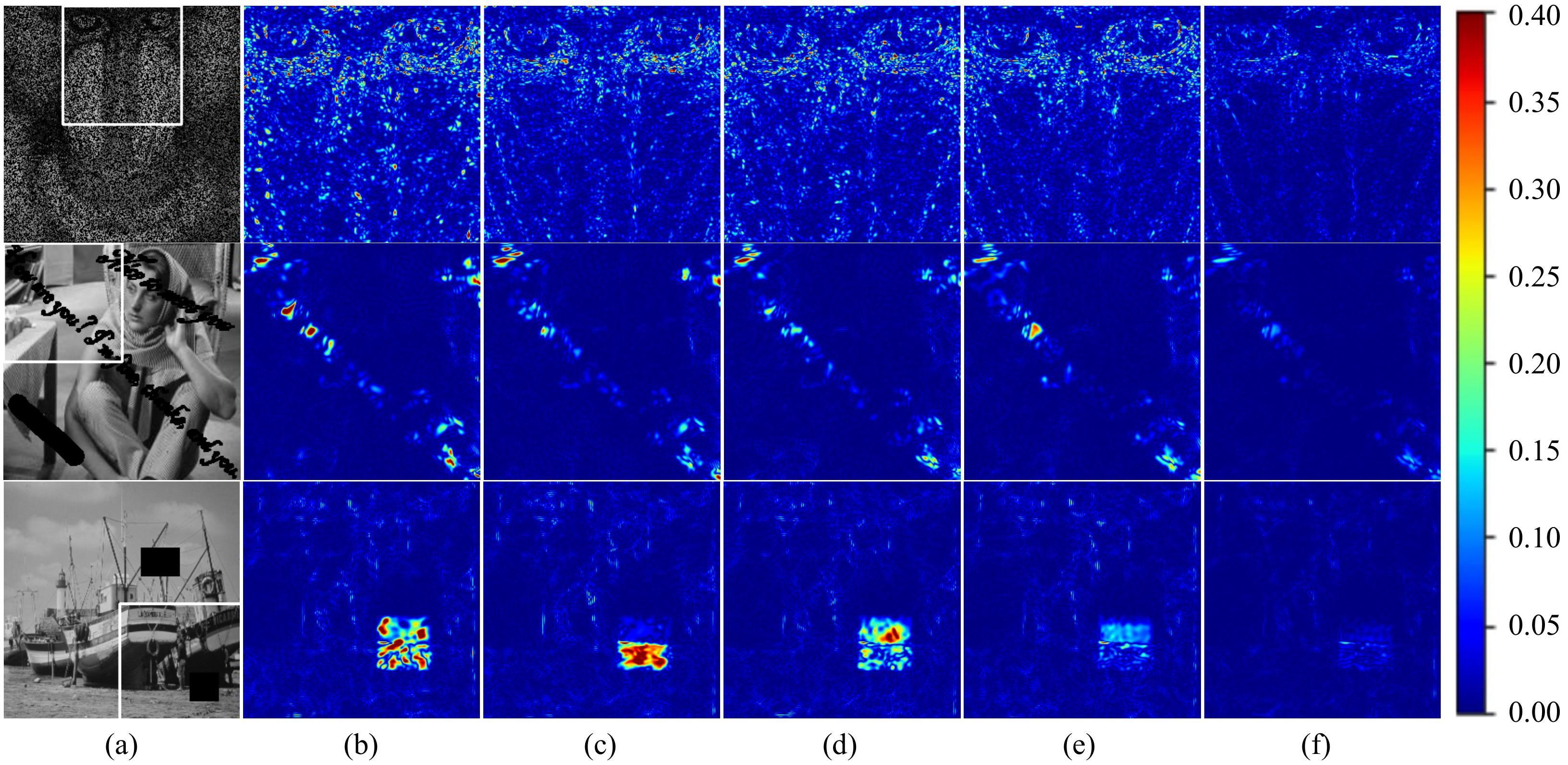}
	\caption{Residual of image inpainting, i.e., $\left|\mathbf{X}-\mathbf{X}^*\right|$ with three types of missing data by different regularized INR including (b) INR without regularization, (c) with TV, (d) $L_2$, (e) AIR, and (f) INRR. The hyper-parameters of benchmark models and algorithms are adopted from the original paper.}
	\label{fig:zoom}
\end{figure*}
\begin{table*}[ht]
	\centering
	\tabcolsep=0.15cm
	\caption{PSNR of recovered images by five data representation models without regularization and with INRR under mixture missing. The hyper-parameters of benchmark models and algorithms are adopted from the original paper. The value in parentheses represents the increment of the PSNR value after using the INRR.}
	\begin{tabular}{lllllllllll}
		\hline
		& DMF & \textbf{DMF+} & ReLU & \textbf{ReLU+} & Siren & \textbf{Siren+} & Garbor & \textbf{Garbor+} & Fourier & \textbf{Fourier+} \\ \hline
		Baboon & 8.2 & \textbf{22.1 (13.9)} & 20.9 & \textbf{21.4 (0.6)} & 17.6 & \textbf{22.7 (5.1)} & 17.5 & \textbf{22.8 (5.4)} & 14.3 & \textbf{22.8 (8.5)} \\
		Man & 8.4 & \textbf{21.7 (13.3)} & 22.2 & \textbf{22.5 (0.3)} & 18.9 & \textbf{22.8 (3.8)} & 17.9 & \textbf{22.0 (4.1)} & 15.3 & \textbf{22.1 (6.9)} \\
		Barbara & 8.8 & \textbf{25.0 (16.3)} & 25.2 & \textbf{26.4 (1.2)} & 23.3 & \textbf{25.6 (2.3)} & 19.4 & \textbf{25.0 (5.5)} & 17.4 & \textbf{25.0 (7.7)} \\
		Boats & 10.9 & \textbf{24.4 (13.4)} & 24.0 & \textbf{25.2 (1.2)} & 22.2 & \textbf{25.7 (3.4)} & 17.6 & \textbf{24.6 (7.0)} & 14.9 & \textbf{24.6 (9.7)} \\
		Cameraman & 6.9 & \textbf{24.5 (17.6)} & 25.7 & \textbf{25.8 (0.1)} & 23.9 & \textbf{25.7 (1.8)} & 18.2 & \textbf{24.7 (6.5)} & 15.2 & \textbf{24.7 (9.4)} \\ \hline
	\end{tabular}
	\label{tab:enhance}
\end{table*}

\textbf{Data types and missing patterns.} 
We consider five gray-scale benchmark images of size $m\times n = 256\times 256$, including Baboon, Man, Barbara, Boats, and Cameraman. Moreover, we study matrix completion with three different missing patterns: random missing, patch missing, and textural missing, which is shown as different parts in \Figref{fig:structure}(a). The default missing rate is $50\%$.

\textbf{Network settings.}
In this section, the INR defaults to SIREN when not otherwise specified \cite{VincentSitzmann2020ImplicitNR}. The INR network is organized in five hidden layers SIREN whose widths are all the same as 256. As to INRR, five hidden layers SIREN is chosen with the same width 32, and the output dimension $r=\max(m,n)$. We use Adam with default settings in \cite{Kingma2015AdamAM} to train all the networks. 

\textbf{Peered methods}
The peered methods include
\begin{enumerate}
	\setlength{\itemsep}{-1pt}
	\item TV: $\mathcal{R}_{\text{TV}}=\sum_{\mathbf{x}_i=(x_i,y_i)\in \mathcal{G}}\left\|\nabla_{\mathbf{x}} \phi_{\boldsymbol{\theta}}(\mathbf{x}_i)\right\|_1$, $\mathcal{G}=\left\{\frac{1}{m},\ldots,\frac{m}{m}\right\}\times \left\{\frac{1}{n},\ldots,\frac{n}{n}\right\}$, which is the discrete version on $\mathcal{G}$.
	\item $L_2$: $\mathcal{R}_{L_2}=\sum_{\ell=1}^{L+1}\left\|\mathbf{W}^{(\ell)}\right\|_2$ which is a common regularizer which is used to regularize NN.
	\item INR-Z: Combining the neighbor of the input with coordinate $[x,y,f(N(x,y))]\in\mathbb{R}^{(N_0+2)\times 1}$ as the input of a new INR $h(\cdot):\mathbb{R}^{N_0+2}\mapsto \mathbb{R}$ as \Figref{fig:structure}(c) shows.
	\item AIR: Adaptive and implicit regularization \cite{ZheminLi2022AdaptiveAI}
	\item INRR: Implicit neural representation regularization proposed in this paper.
\end{enumerate}

\subsection{Image representation with various missing patterns}

We apply INRR for matrix completion (or image inpainting) on three types of missing patterns. A few related models are also used for comparison.

\textbf{Adaptive to training data}. We compare vanilla INRR with several improved models in the following experiments, including TV, $L_2$, AIR, INRR, and INR-Z. \Tabref{tab:diffreg} lists the PSNRs of recovered images using the aforementioned improved models for different data with different missing patterns. The results show that the non-local regularization methods, including AIR and INRR, significantly outperform the vanilla INR. Furthermore, INRR is much better than AIR since INRR integrates the smoothness of Laplacian matrix into the DE regularizer. The residual of recovered images $\left|\mathbf{X}-\mathbf{X}^*\right|$ corresponding to \Tabref{tab:diffreg} are shown in \Figref{fig:zoom}. Unlike other INR-regularized methods that perform well for random missing cases but poorly for other missing patterns, INRR consistently gives visually appealing results. To conclude, INRR achieves decent results qualitatively and quantitatively independent of sampling mode of training data.

\textbf{Adaptive to data representation}. 
To distinguish the effect of INRR regularizer from the vanilla INR model, \Tabref{tab:enhance} lists the PSNRs of recovered images by several data representations which INRR regularizes. The data representation includes deep matrix factorization (DMF) \cite{SanjeevArora2019ImplicitRI}, FCN with ReLU activation function, SIREN \cite{VincentSitzmann2020ImplicitNR}, the filter neural network with Gabor and Fourier filter, respectively \cite{Fathony2021MultiplicativeFN}.
The INRR regularized models are denoted by '.+' in \Tabref{tab:enhance}. The mixture missing pattern is shown in \Figref{fig:Whenbad}(a). The results shown in \Tabref{tab:enhance} and \Figref{fig:Whenbad}(b)(c) both illustrate that INRR significantly improves the performance of recently proposed data representation methods without regularization. Overall, INRR is a general regularizer not limited to being combined with a particular data representation model.

\section{Why INRR performs better}
Now we have shown that INRR achieves excellent performance in image representation (image inpainting as an example) under different missing patterns. In this section, the reasons why INRR performs better than other peered methods are analyzed carefully. Firstly, the smoothness of $\mathbf{L}$ learned by INRR is demonstrated by experiments. Then a heuristic connection between INRR, implicit bias, and the momentum method is built.

\subsection{Tiny INR smooths Laplacian matrix implicitly}
\begin{figure}[htb]
	\begin{center}
		\begin{tabular}{cc}
			\hspace{-0.3cm}\includegraphics[width=0.48\columnwidth]{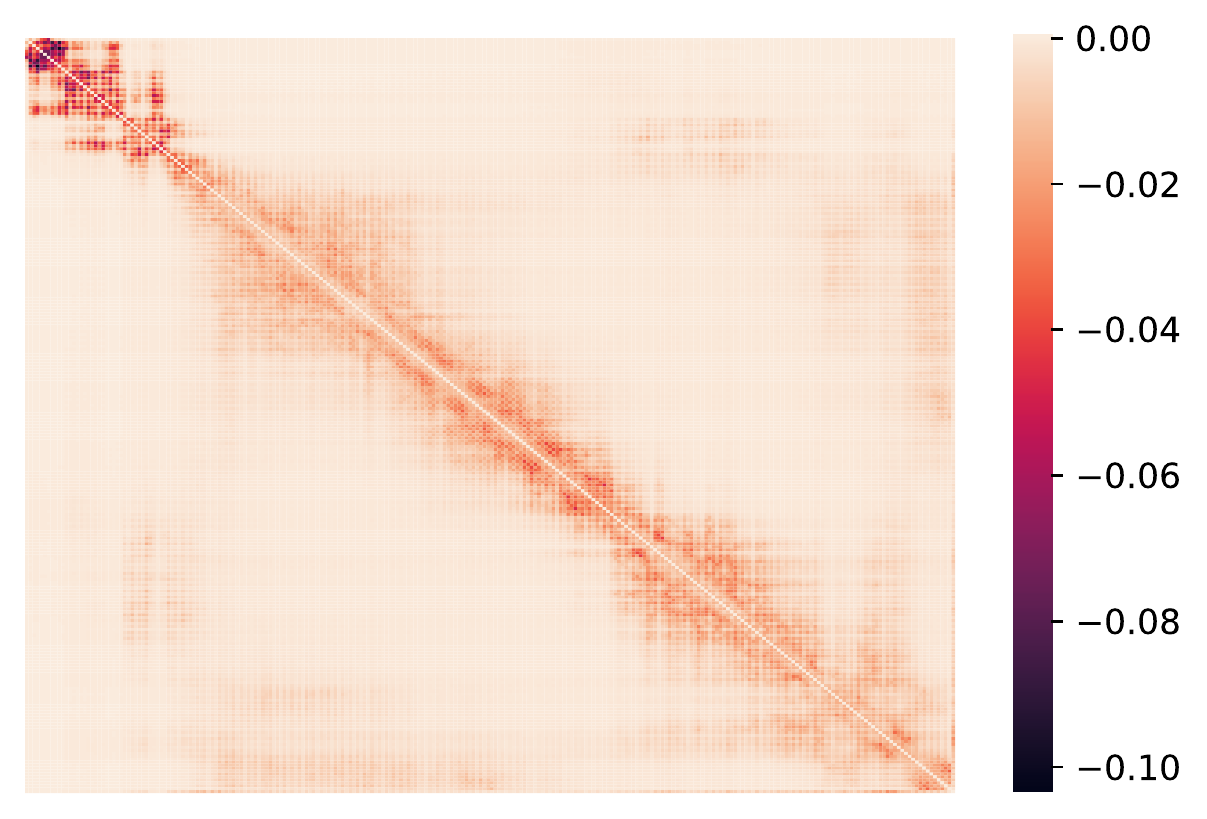}&
			\hspace{-0.3cm}\includegraphics[width=0.48\columnwidth]{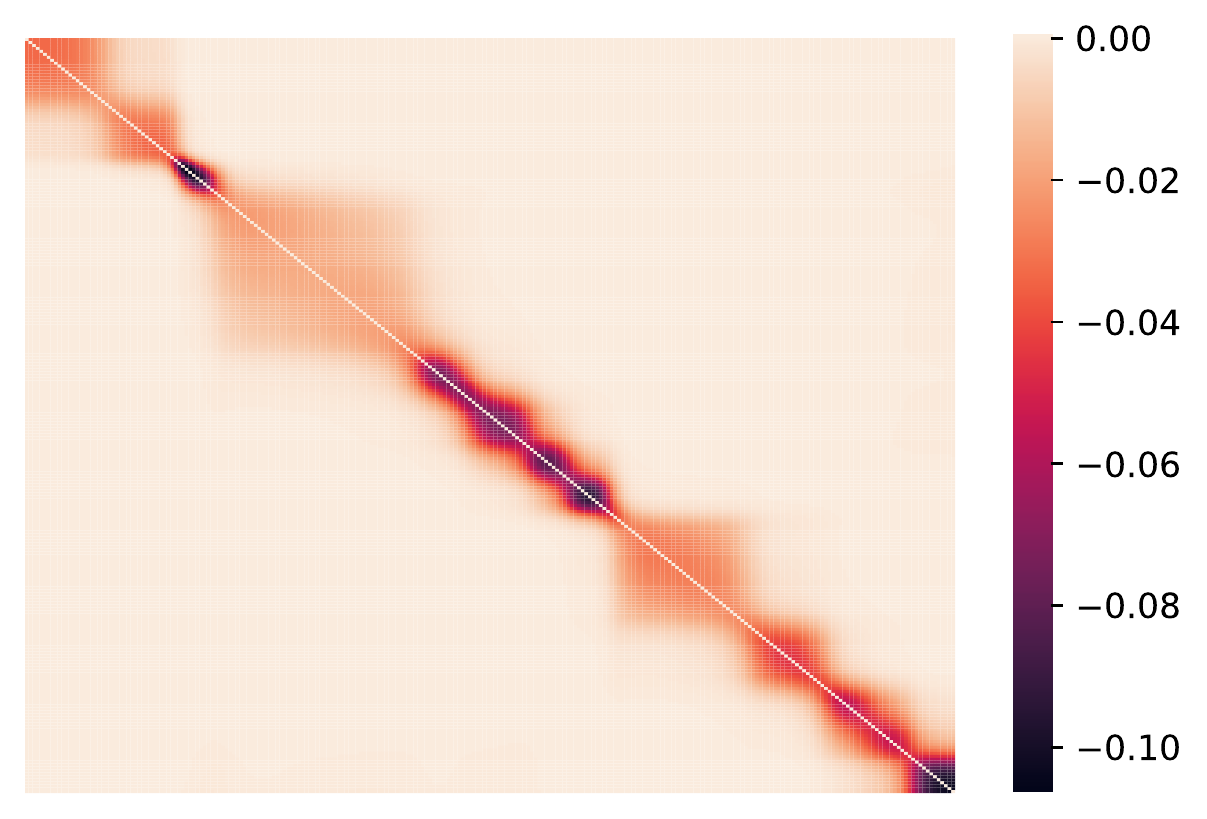}\\
			{\hspace{-0.3cm}\footnotesize (a) AIR} & \hspace{-0.3cm}{\footnotesize (b) INRR} 
		\end{tabular}
	\end{center}\vspace{-0.4cm}
	\caption{$\mathbf{L}$ at $t=2000$ learned by (a) AIR and (b) INRR, respectively. All the methods are based on the SIREN to super-resolution Baboon, which is down-sampled from $256\times 256$ to $128\times 128$.}\label{fig:smoothness}
\end{figure}

Parameterizing DE with a tiny INR is the key of INRR. In this section, we focus on illustrating the benefit of this parameterization. A $256\times 256$ Baboon is down-sampled to $128\times 128$, and then AIR and INRR are used to regularize INR to recover the original image based on the sampled data. \Figref{fig:smoothness} shows the Laplacian matrix $\mathbf{L}$ learned by AIR and INRR, respectively. The $\mathbf{L}$ learned by AIR (\Figref{fig:smoothness}(a)) 
is discontinuous with high probability at those locations that are not sampled, while the $\mathbf{L}$ learned by INRR (\Figref{fig:smoothness}(b)) is much more continuous. The continuous $\mathbf{L}$ introduced by the tiny INR is more consistent with practice. 

\subsection{INRR behaves like a momentum}
\begin{figure*}[htb]
	\begin{center}
		\begin{tabular}{ccc}
			\hspace{-0.2cm}\includegraphics[width=0.66\columnwidth]{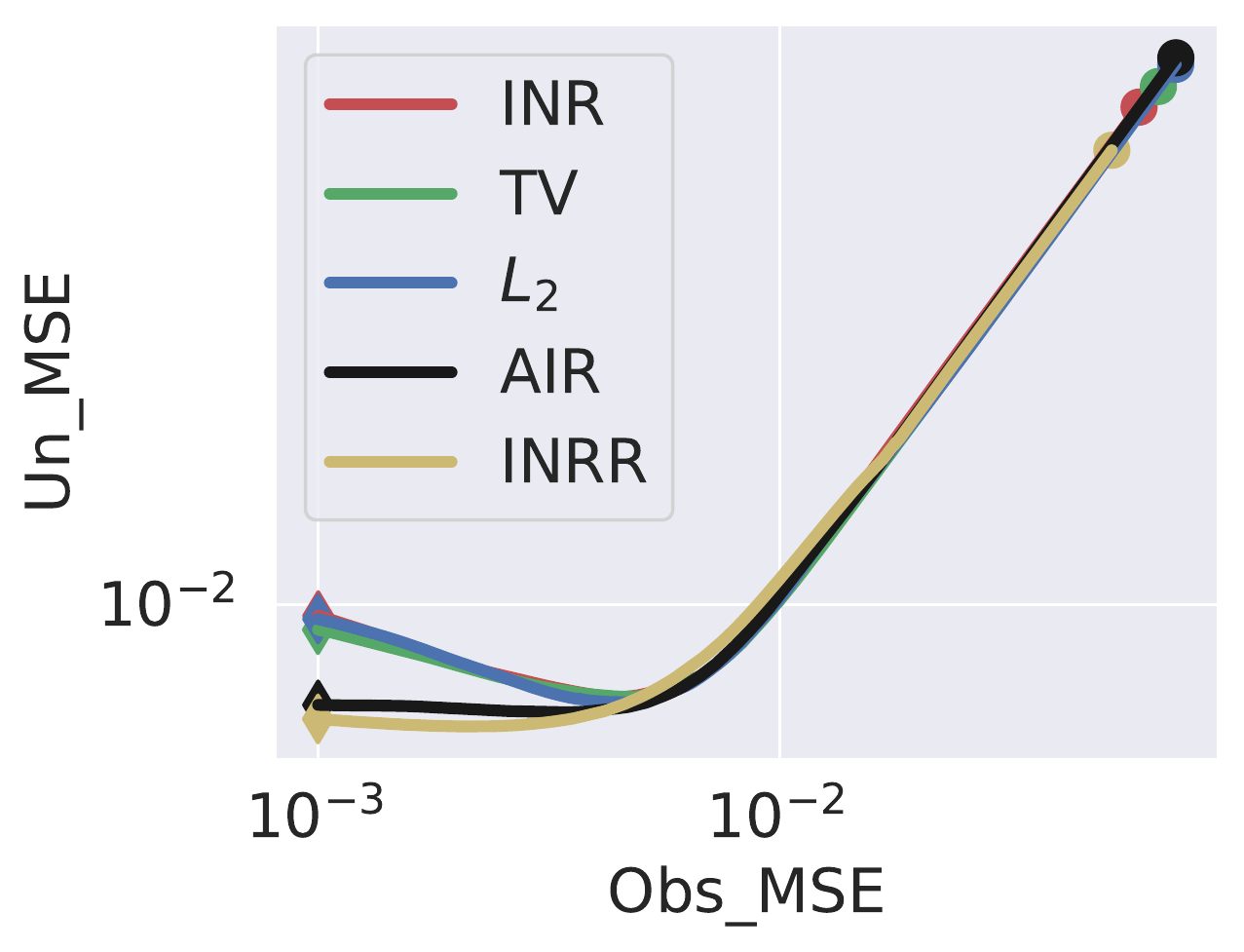}&
			\hspace{-0.3cm}\includegraphics[width=0.7\columnwidth]{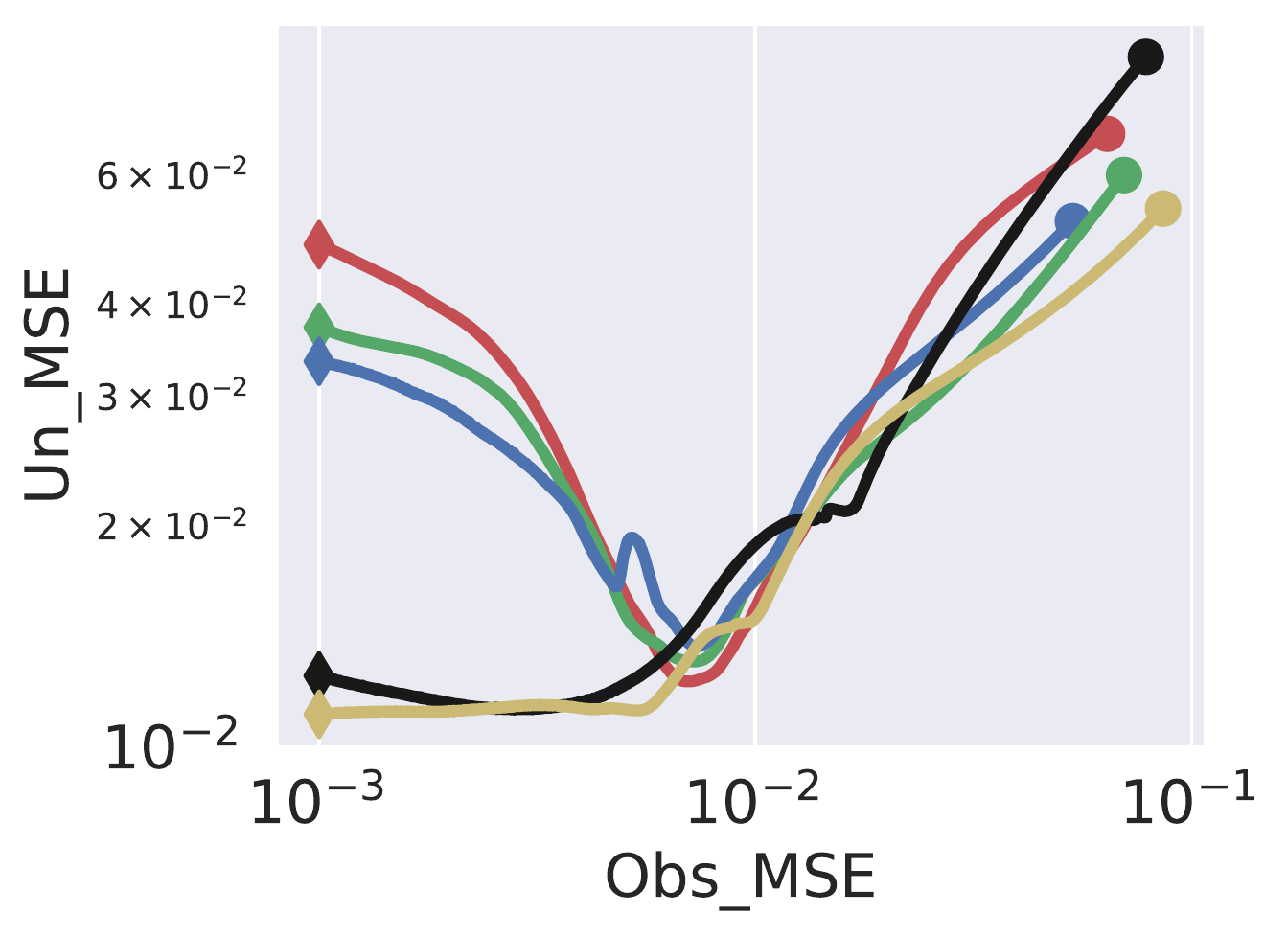}&
			\hspace{-0.3cm}\includegraphics[width=0.66\columnwidth]{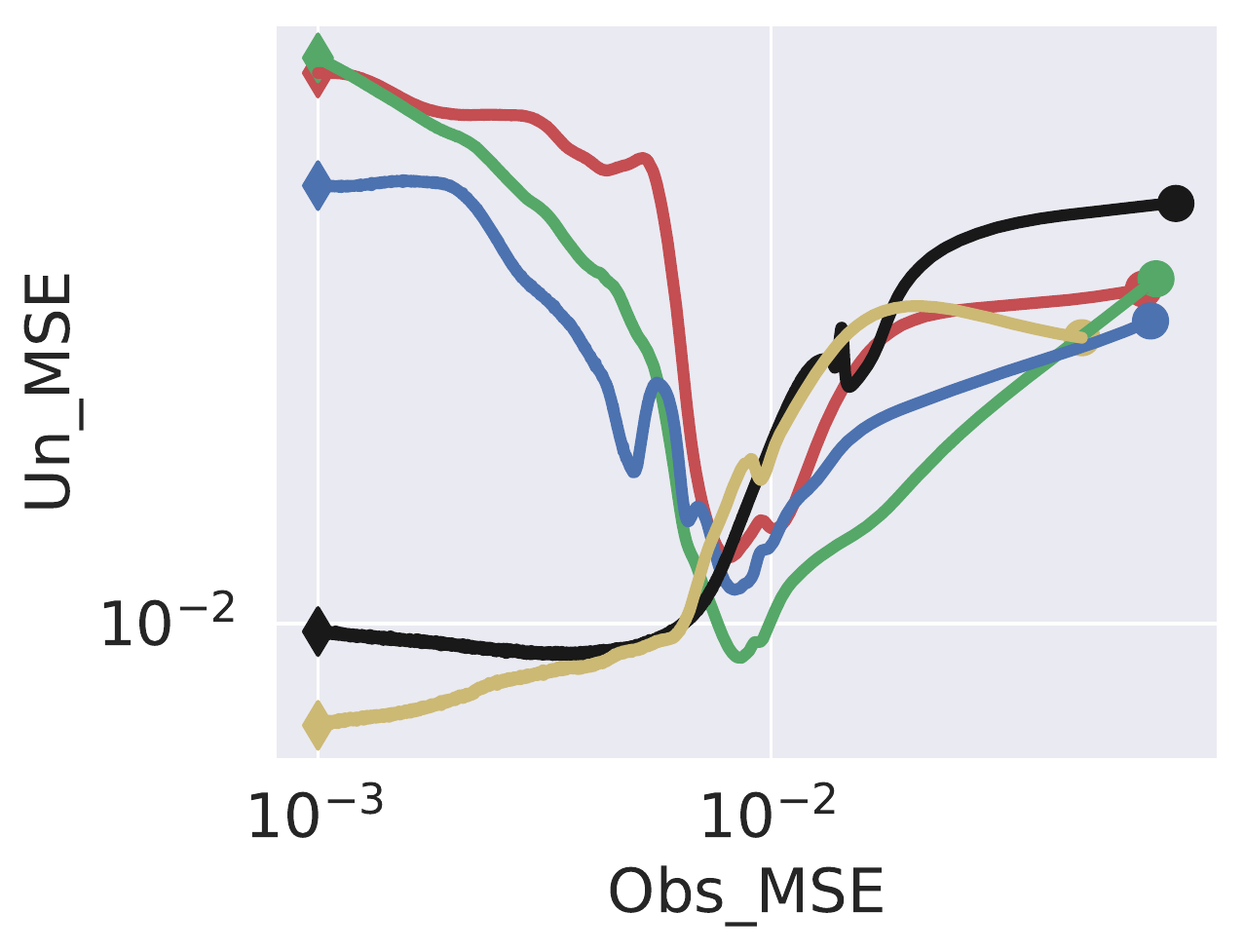}\\
			{\footnotesize (a) Random} & \hspace{-0.3cm} {\footnotesize (b) Textural} &\hspace{-0.3cm}  {\footnotesize (c) Patch} \\
		\end{tabular}
	\end{center}\vspace{-0.3cm}
	\caption{Trajectories of training INR model without regularization, with TV, $L_2$, AIR and INRR regularization for inpainting the Baboon image with three types of missing types of missing pixels: (a) randomly missing $50\%$, (b) textural missing, and (c) patch missing. The dot point indicates $\mathbf{X}(0)$ and the diamond shape of different color indicate the MSE on training data achieve $10^{-3}$.}\label{fig:tra}
\end{figure*}
We connect INRR with the momentum method in this subsection. As \Figref{fig:heatmap} shows, INRR tends to vanish during training. Then INR with INRR converges to the vanilla INR model. First, compare INRs with and without INRR by the optimization trajectory. In \Figref{fig:tra}, we plot the MSE's trajectory during training. At the beginning of training, the observed and unobserved MSEs of the five models drop similarly. However, these five models perform dramatically differently near the convergence. When the observed MSE becomes smaller, the model learns details in observed elements. The unobserved MSE increased during the observed MSE decrease in the vanilla INR, INR+TV, and INR+$L_2$ cases; we name this phenomenon over-fitting. INRR and AIR keep the decaying trend for both observed and unobserved MSEs. Significantly, the proposed INRR keeps the decaying trend better than AIR due to the extra smoothness introduced by a tiny INR.

Looking back into the training process of INRR, the update of $\mathbf{X}(t+1)$ involves both $\mathbf{X}(t)$ and $\mathbf{L}(t)$, and the update of $\mathbf{L}(t)$ depends on $\mathbf{X}(t-1)$. To understand the training dynamics, we consider the following simplified model:
$$
\Min_{\mathbf{X},\mathbf{L}} \left\{\mathcal{L}+\lambda tr(\mathbf{X}^\top \mathbf{L}\mathbf{X})\right\},
$$
and we have
$$
\nabla_{\mathbf{X}(t)}\mathbf{L}=\nabla_{\mathbf{X}(t)}\mathcal{L} +2\lambda\mathbf{L}(t)\mathbf{X}(t),
$$
where $\mathcal{L}$ is the fidelity term, $\mathbf{L}(t)$ is the function of $\left\{\mathbf{X}(t_0)\mid t_0<t\right\}$ as $\mathbf{L}(t)$ is updated based on $\mathbf{X}(t-1)$. Therefore, every iteration step of INRR leverages all the previously learned information $\left\{\mathbf{X}(t_0)\mid t_0<t\right\}$. Note that the update of both vanilla INR, INR+TV, and INR+$L_2$ only depend on $\mathbf{X}(t)$. From this viewpoint, INRR shares a similar spirit as the momentum method, which leverages history to improve performance.

\subsection{INRR connects implicit bias with multi-scale self-similarity}
\begin{figure*}[htb]
	\centering
	\includegraphics[width=0.8\linewidth]{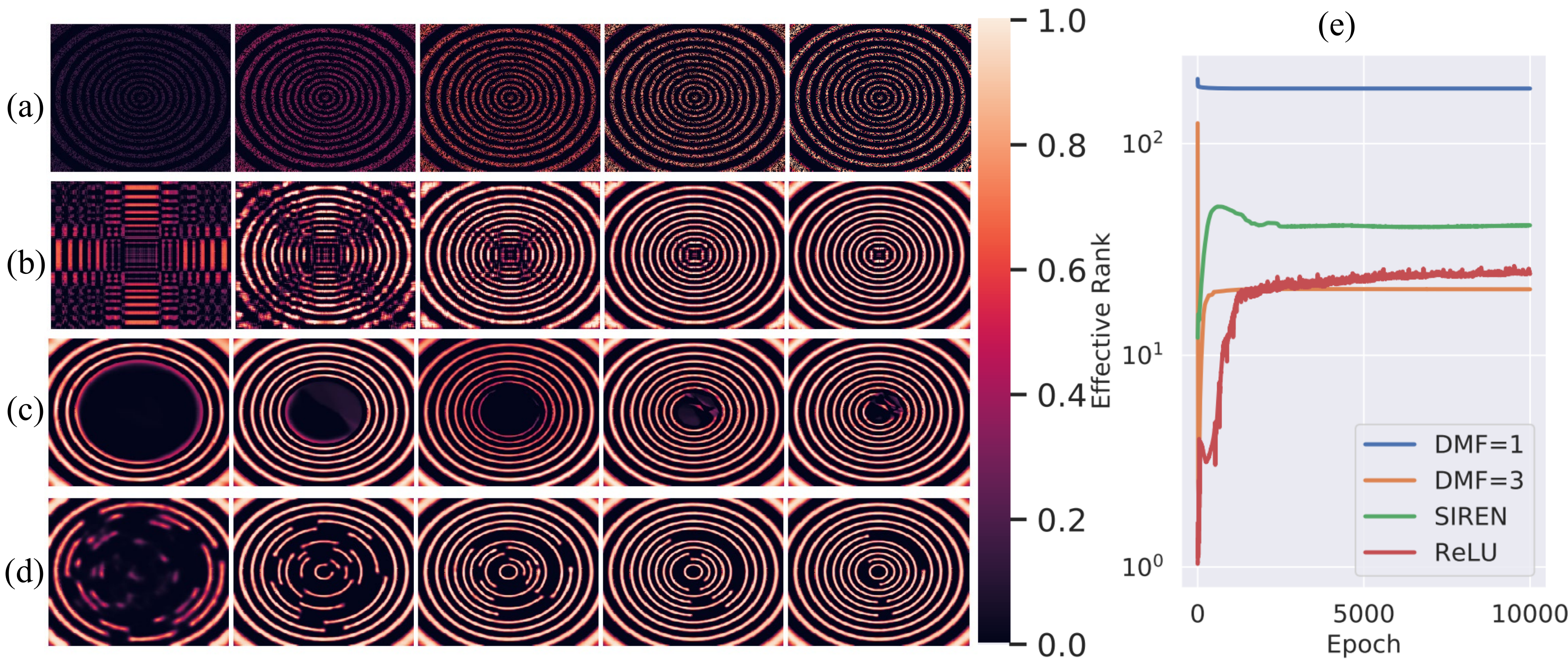}
	\caption{Fitting a $256\times 256$ synthetic data which randomly missing $50\%$ pixels at different training step with (a) single matrix (DMF with only one factor), (b) DMF with three factors, (c) ReLU neural network, and (d) SIREN. The effective rank of the fitted matrix is shown in (e).}
	\label{fig:bias}
\end{figure*}

\begin{figure*}[htb]
	\centering
	\includegraphics[width=0.8\linewidth]{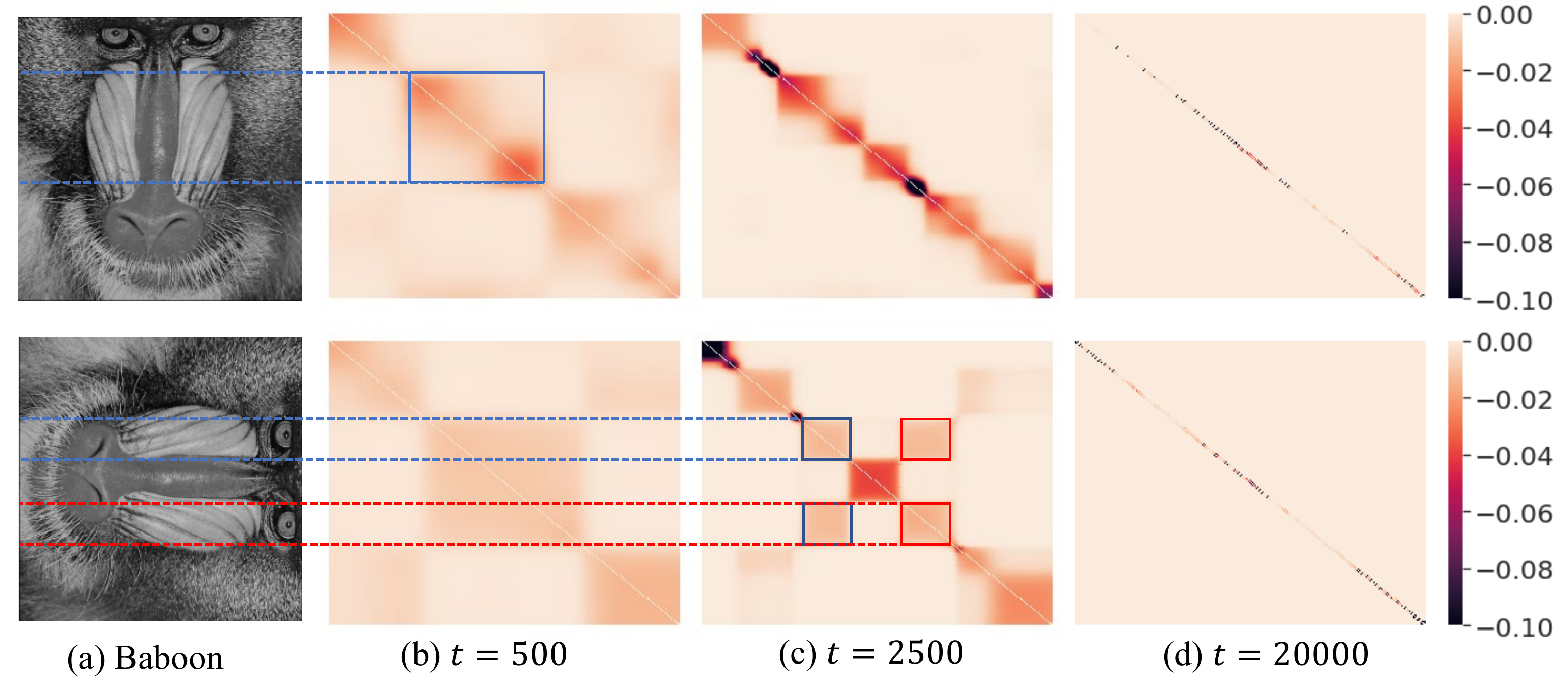}
	\caption{Learned $\mathbf{L}_r(t)$ and $\mathbf{L}_c(t)$ during training. (a): first and second rows depict the Baboon image and its rotation. (b)-(d): first/second row shows the heatmap of $\mathbf{L}_r/\mathbf{L}_c$ at different $t$. A darker color indicates a stronger similarity captured by the adaptive regularizer. The $(i,j)$-th element in the heatmap of $\mathbf{L}_r(t)$ has a darker color than the $i,j'$-th element indicates that the $i$-th row is more related to the $j$-th row compared with the $j'$-th row.}
	\label{fig:heatmap}
\end{figure*}

\begin{figure}[htb]
	\begin{center}
		\begin{tabular}{c}
			\vspace{-0.2cm}\includegraphics[width=0.9\columnwidth]{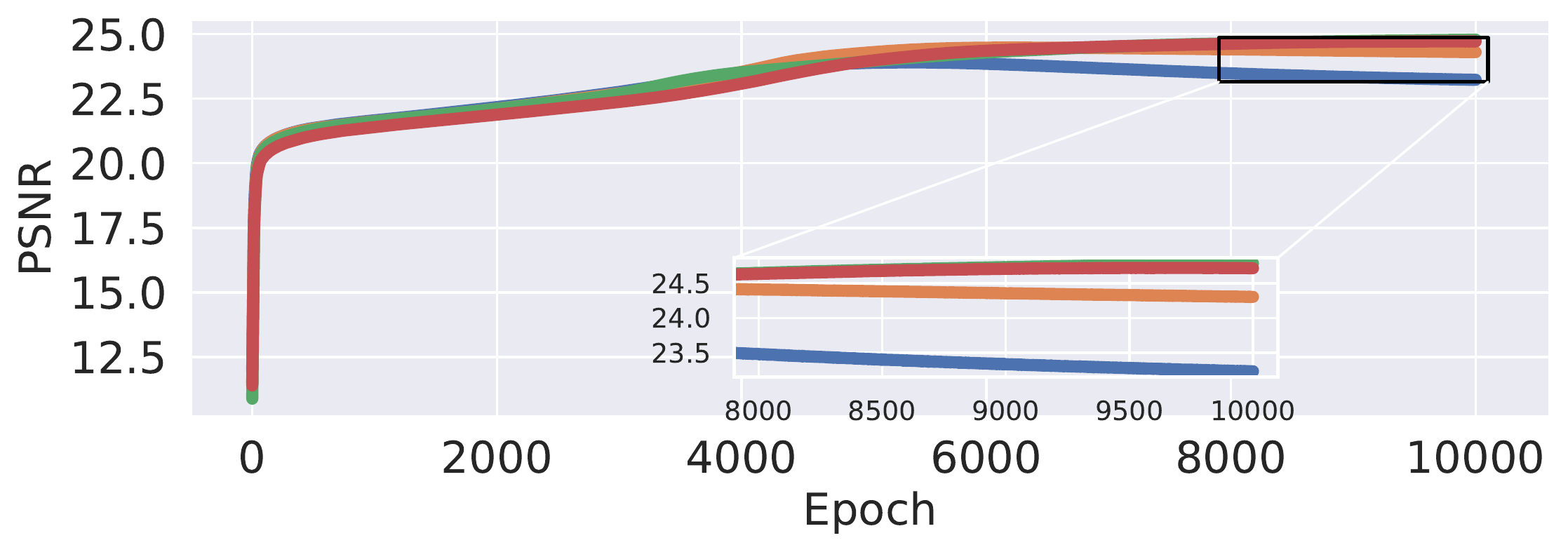}\\
			{\footnotesize (a) Random}\\
			\vspace{-0.2cm}\includegraphics[width=0.9\columnwidth]{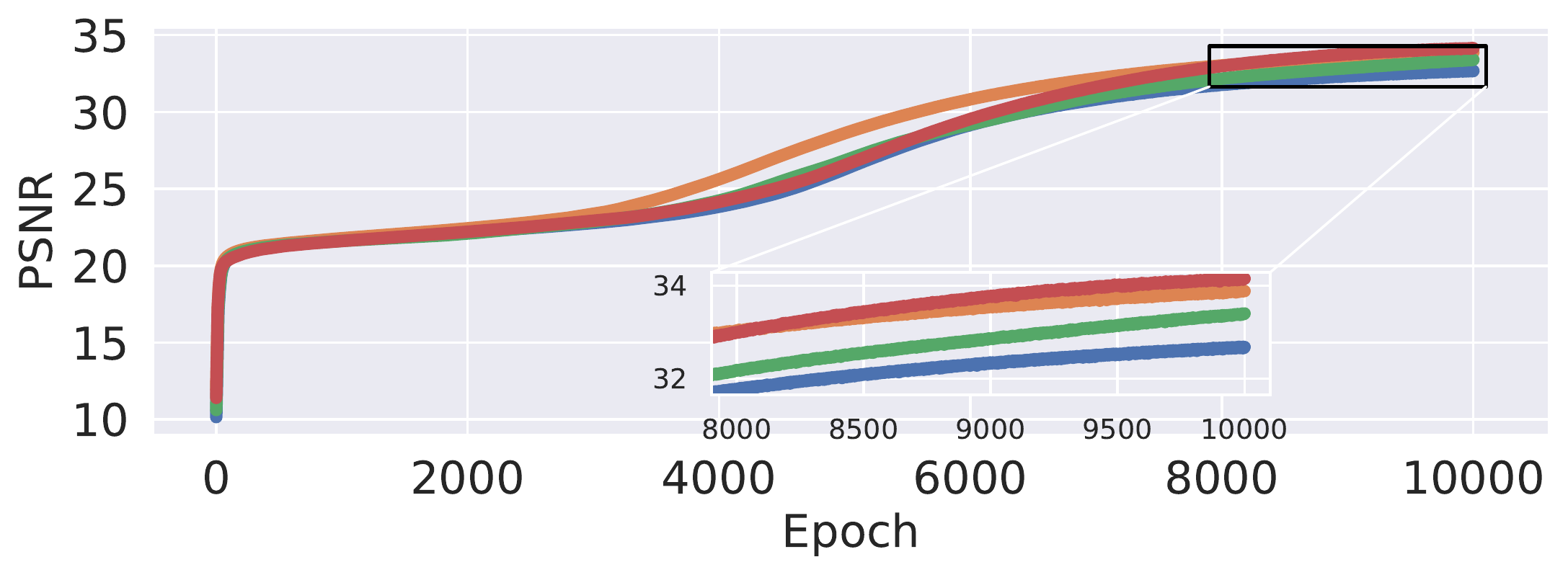}\\
			{\footnotesize (b) Patch}\\
			\vspace{-0.2cm}\includegraphics[width=0.9\columnwidth]{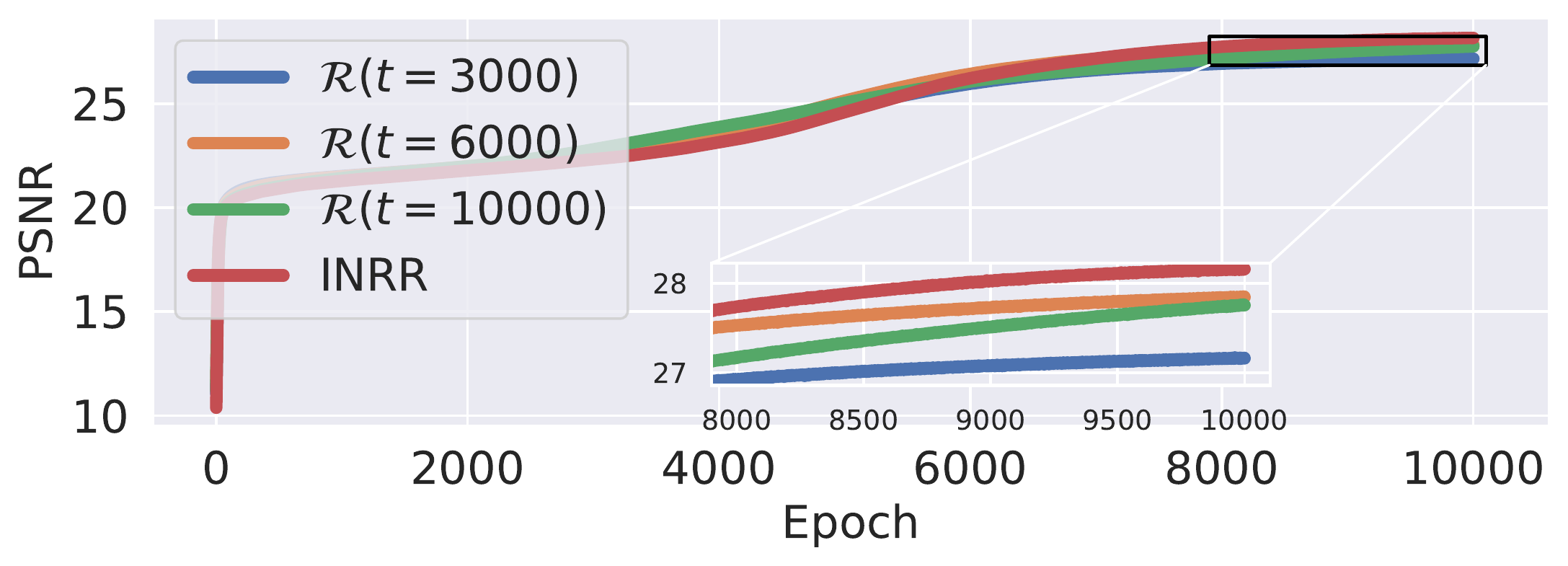}\\
			{\footnotesize (c) Textural} 
		\end{tabular}
	\end{center}\vspace{-0.7cm}
	\caption{Contrasting the adaptive regularizer with the fixed regularizer for Baboon image recovery. We consider inpainting the Baboon image with three types of missing pixels: (a) randomly missing $50\%$ pixels, (b) patch missing, and (c) textural missing. The red lines plot the PSNR during the training of the vanilla INRR. The remaining three lines in each figure indicate replacing $\mathbf{L}_r$ and $\mathbf{L}_c$ with $\mathbf{L}_r(t)$ and $\mathbf{L}_c(t)$ at $t=3000,6000$ and $10000$, respectively.}
	\label{fig:diff_epoch}
\end{figure}

\textbf{Implicit bias of NN.}
We then demonstrate other properties of INRR by connecting implicit bias with multi-scale self-similarity. The implicit bias of NN is used to explain the generalization ability of NN in recent years \cite{ZhiQinJohnXu2019FrequencyPF,TaoLuo2021TheoryOT,YuanCao2021TowardsUT,AbdulkadirCanatar2021SpectralBA,SanjeevArora2019ImplicitRI,TomasVakeviius2019ImplicitRF,PengZhao2019ImplicitRV}. As \Figref{fig:bias} shows, we fit synthetic data with DMF with one factor, DMF with three factors, ReLU FCN, and SIREN, respectively. The synthetic data is sampled from function $s(x,y)=\sin\left(25\pi\sin\left(\frac{\pi}{3}\cdot \sqrt{x^2+y^2}\right)\right)$, where $\{(x_i,y_j)|i,j\}$ is a uniform $256\times 256$ grid on $[-1,1]\times [-1,1]$, where the local frequency of the synthetic data increases from boundary to center. All the networks except DMF with one factor evolve from a low complexity pattern to a high complexity one. ReLU FCN and SIREN first fit the low-frequency components and then gradually fit the high-frequency components \cite{ZhiQinJohnXu2019FrequencyPF,TaoLuo2021TheoryOT,YuanCao2021TowardsUT,AbdulkadirCanatar2021SpectralBA}.  More specifically, the effective rank of DMF with three factors, SIREN, and ReLU FCN, increases gradually as the line plot, where the effective rank can measure the effective dimension of the matrix with more accuracy than discrete rank \cite{OlivierRoy2007TheER,SanjeevArora2019ImplicitRI}. 

\textbf{Multi-scale similarity captured by INRR.}
Then we turn to explain the multi-scale similarity seized by INRR. Due to the implicit bias of fidelity term, INRR can capture different scales of data similarity. The heatmaps of Laplacian matrices $\mathbf{L}_r(t)$ and $\mathbf{L}_c(t)$ for Baboon are shown in \Figref{fig:heatmap}. A few large blocks appear in $\mathbf{L}_r(500)$ and $\mathbf{L}_c(500)$ \Figref{fig:heatmap}(b), which reflect the similarity in a large scale. Then the size of blocks becomes smaller while the number of blocks increases at $t=2500$ in \Figref{fig:heatmap}(c), which reflects the similarity on a smaller scale. The values in these blocks reflect the substantial similarity of the corresponding highlighted patches of the original Baboon, which echoes our intuition. Moreover, as the training goes further, both $\mathbf{L}_r(t)$ and $\mathbf{L}_c(t)$ 
focus on reflecting the similarity of the neighbor at $t=20000$ in \Figref{fig:heatmap}(d), which is similar to the TV.

\textbf{The importance of learned INRR.}
The results confirm that INRR captures the similarity from large to small. Next, we experimentally illustrate that the learned $\mathbf{L}_c$ and $\mathbf{L}_c$ by INRR are crucial for image representation. Fix $\mathbf{L}_r$ and $\mathbf{L}_c$ at a specific training step for INRR, and then compare INRR with the overall adaptive $\mathbf{L}_r$ and $\mathbf{L}_c$.

We contrast the vanilla INRR and INRR with fixed Laplacian matrices (let $t=3000,6000$ and $10000$ respectively) for Baboon image inpainting. \Figref{fig:diff_epoch} shows how the PSNR changes during training. INRR, which continuously updates the regularization during training, performs best for all missing patterns. Fixing Laplacian matrices helps reduce the computation costs during training. However, as the optimal $t^*$ is varied with missing patterns, the learned Laplacian matrices are more applicable.

\section{Conclusion}
\label{sec:conclusion}
This paper proposes a novel regularizer named INRR, which significantly improves INR's representation performance, especially when the training data is sampled arbitrarily. INRR parameterizes the Laplacian matrix in DE by a tiny INR and then adaptively learns the non-local similarities hidden in image data. INRR is a generic framework for integrating multiple prior into a single regularizer, decreasing the redundancy of the regularizer. The connection among INRR, momentum term, implicit bias, and multi-scale self-similarity deserve further theoretical analysis.

{\small
\bibliographystyle{ieee_fullname}
\bibliography{11_references}
}

\ifarxiv \clearpage 
\appendix
\label{sec:appendix}

\section{Theoretical analysis}
\subsection{Fourier feature map induce a shift-invariant kernel regression}
\label{subsec:app_shift}
Notice that $\phi_{\text{NTK}}(\mathbf{x})$ should be shift-invariant, i.e., if we shift the training data $\left\{(\mathbf{x}_i, z_i)\right\}_{i=1}^N$ to $\left\{(\mathbf{x}_i+\Delta \mathbf{x}, z_i)\right\}_{i=1}^N$, and the corresponding kernel regression is $\phi_{\text{NTK}}^+(\mathbf{x})$, we look forward $\phi_{\text{NTK}}^+(\mathbf{x}+\Delta \mathbf{x})=\phi_{\text{NTK}}(\mathbf{x})$. Researchers encode the shift-invariant property by a Fourier feature map $\gamma(\mathbf{x})=\frac{1}{\sqrt{D}}[\cos \mathbf{Bx}^\top,\sin \mathbf{Bx}^\top]^\top:\mathbb{R}^d\mapsto \mathbb{R}^{2D}$ as input, where $\mathbf{x}\in \mathbb{R}^{d}$, $\mathbf{B}\in \mathbb{R}^{D\times d}$, and $\mathbf{B}_{ij}\sim \mathcal{N}(0,\delta)$ \cite{MatthewTancik2020FourierFL}. The NTK can be written as $h_{\text{NTK}}(\mathbf{x}_i^\top\mathbf{x}_j)$, $h_{\text{NTK}}:\mathbb{R}\mapsto \mathbb{R}$ when $\mathbf{x}_i$ on a hypersphere, so NTK with feature map can be composed as $h_{\text{NTK}}\left(\gamma(\mathbf{x}_i)^\top \gamma(\mathbf{x}_j)\right)=h_{\text{NTK}}\left(\frac{1}{D}\mathbf{1}_D^\top\cos\left(\mathbf{B}(\mathbf{x}_i-\mathbf{x}_j)\right)\right)$ which is shift-invariant. Then $\phi_{\text{NTK}}(\gamma(\mathbf{x}+\Delta \mathbf{x}))=\phi_{\text{NTK}}(\gamma(\mathbf{x}))$, and
$$
\phi_{\text{NTK}}'(\mathbf{x})=\phi_{\text{NTK}}(\gamma(\mathbf{x}))=\sum_{i=1}^N (\mathbf{H}^{-1}\mathbf{z})_i h_{\text{NTK}}(\mathbf{x}_i,\mathbf{x}),
$$
where $\mathbf{H}$ is an $n\times n$ PSD matrix with entries $\mathbf{H}_{ij}=h_{\text{NTK}}(\mathbf{x}_i,\mathbf{x}_j)$.

\subsection{Proof of main theorem}
\begin{proof}[Proof of Theorem 1]
	$$
	\begin{aligned}
	k_D(\mathbf{x}_i,\mathbf{x}_j)
	&=h_{\text{NTK}}\left(\frac{1}{D}\mathbf{1}_D^\top\cos\left(\mathbf{B}(\mathbf{x}_i-\mathbf{x}_j)\right)\right)\\
	&=h_{\text{NTK}}\left(\sum_{l=1}^D \frac{1}{D}\cos\left(\mathbf{B}_{l,:}(\mathbf{x}_i-\mathbf{x}_j)\right)\right),\\
	\end{aligned}
	$$
	where $\mB_{l,:}$ is the $l$-th row of $\mB$.
	
	Therefore, 
	$$
	\lim_{D\rightarrow \infty}k_D(\mathbf{x}_i,\mathbf{x}_j)= h_{\text{NTK}}\left(\mathbb{E}_{\mathbf{b}_{l}\sim \mathcal{N}(0,\delta)}\cos\left(\mathbf{b}^\top(\mathbf{x}_i-\mathbf{x}_j)\right)\right),
	$$
	where $\mathbf{b}\in\mathbb{R}^d$. Furthermore, $\lim_{D\rightarrow \infty}k_D(\mathbf{x}_i,\mathbf{x}_j)=h_{\text{NTK}}(e^{-\delta^2 \left\|\mathbf{x}_i-\mathbf{x}_j\right\|_2^2})$.
	$\hfill\blacksquare$ 
\end{proof}

\begin{proof}[Proof of Corollary 1]
	As 
	$$
	\lim_{\substack{\delta\rightarrow \infty\\D\rightarrow \infty}}k_D(\mathbf{x},\mathbf{X})=\left\{ \begin{array}{cc}
	h(1)\mathbf{e}_i^\top&\mathbf{x}\in\left\{\mathbf{x}_i\right\}_{i=1}^N,\\
	h(0)\boldsymbol{1}_N^\top&\mathbf{x}\not\in\left\{\mathbf{x}_i\right\}_{i=1}^N,
	\end{array}\right.
	$$
	then,
	$$
	\lim_{\substack{\delta\rightarrow \infty\\D\rightarrow \infty}}k_D(\mathbf{X},\mathbf{X})=h(0)\boldsymbol{1}_N\boldsymbol{1}_N^\top+\left(h(1)-h(0)\right)\mI_N.
	$$
	That is, the singular value of $\lim_{\substack{\delta\rightarrow \infty\\D\rightarrow \infty}}k_D(\mathbf{X},\mathbf{X})$ are $h(1),h(1)-h(0),\ldots,h(1)-h(0)$. It's evidence that $k_{\infty}(\mathbf{X},\mathbf{X})=\lim_{\substack{\delta\rightarrow \infty\\D\rightarrow \infty}}k_D(\mathbf{X},\mathbf{X})$ is invertible when $h(1)\neq h(0)$ and $h(1)\neq 0$. Then
	$$\Phi_{\text{NTK}}'(\mathbf{X})=k_{\infty}(\mathbf{X},\mathbf{X})k_{\infty}(\mathbf{X},\mathbf{X})^{-1}\mathbf{z}=\mathbf{z},$$
	that is $\phi_{\text{NTK}}'(\mathbf{x}_l)=z_l$.
	
	As for $\mathbf{x}\not\in \left\{\mathbf{x}_i\right\}_{i=1}^N$,
	\begin{equation}
		\label{eq:phiinf}
		\begin{aligned}
			\Phi_{\text{NTK}}'(\mathbf{x})&=k_\infty (\mathbf{x},\mathbf{X})\cdot k^{-1}_{\infty}(\mathbf{X},\mathbf{X})\mathbf{z}\\
			&=h(0)\boldsymbol{1}_N^\top k^{-1}_{\infty}(\mathbf{X},\mathbf{X})\cdot \mathbf{z}.
		\end{aligned}
	\end{equation}
	Note that $k_\infty(\mathbf{X},\mathbf{X})$ is a particular matrix which has same column summation that is 
	$$
	\boldsymbol{1}_N^\top\cdot k_\infty(\mathbf{X},\mathbf{X})=\left((N-1)h(0)+h(1)\right)\boldsymbol{1}_N^\top,
	$$
	therefore its corresponding eigenvalue is $(N-1)h(0)+h(1)$. 
	Furthermore, as
	$$
	\frac{1}{(N-1)h(0)+h(1)}\boldsymbol{1}_N^\top=\boldsymbol{1}_N^\top\cdot k_\infty^{-1}(\mathbf{X},\mathbf{X}),
	$$
	we have, $\boldsymbol{1}_N^\top$ is the left eigenvector of $k_\infty(\mathbf{X},\mathbf{X})^{-1}$ and the corresponding eigenvalue is $\frac{1}{(N-1)h(0)+h(1)}$. Then bring it back to Eq.\ref{eq:phiinf}, we have 
	$$
	\Phi_{\text{NTK}}'(\mathbf{x})=\frac{h(0)}{(N-1)h(0)+h(1)}\boldsymbol{1}_N^\top \mathbf{z}.
	$$
	$\hfill\blacksquare$ 
\end{proof}

\section{Explain the proposed method step-by-step}
For simplicity, we focus on a gray-scale image inpainting task to illustrate the workflow of our method.

\noindent \textbf{Task}:
Given a partially observed image $\mathbf{X}$ on $\mathcal{X}$, where $\mathcal{X}\subseteq \mathcal{G}=\left\{(\frac{i}{m},\frac{j}{n})|i\in\{1,2,\cdots,m\},j\in\{1,2,\cdots,n\}\right\}$, i.e., $\mathcal{Z}=\left\{\mathbf{X}_{i,j}\mid (\frac{i}{m},\frac{j}{n})\in \mathcal{X}\right\}$, find $\mathbf{X}$ on unobserved $\mathcal{G}\backslash \mathcal{X}$.

\noindent \textbf{Input}: Training set $\mathcal{X}\times \mathcal{Z}$; initial network parameters $\boldsymbol{\theta}_a(0)=\{\boldsymbol{\theta}(0),\boldsymbol{\theta}_r(0),\boldsymbol{\theta}_c(0)\}$; super-parameters $\lambda_r,\lambda_c$; iteration step $t=0$.

\noindent \textbf{Step 1}: \textit{Calculate loss function.} Loss function is  
$\mathcal{L}_{a}(\boldsymbol{\theta}_a)=\mathcal{L}(\boldsymbol{\theta}(t), \mathcal{X}, \mathcal{Z})+\lambda_r \mathcal{R}(\boldsymbol{\theta}_r(t))+\lambda_c \mathcal{R}(\boldsymbol{\theta}_c(t))$, where $\mathcal{L}=\sum_{(\mathbf{x}_i,z_i)\in\mathcal{X}\times \mathcal{Z}}\left\|\phi_{\boldsymbol{\theta}(t)}(\mathbf{x}_i)-z_i \right\|_2^2$,  $\mathcal{R}(\boldsymbol{\theta}_r(t))=\text{tr} \left(\mathbf{X}^\top \mathbf{L}(\boldsymbol{\theta}_r(t))\mathbf{X}\right)$ and $\mathcal{R}(\boldsymbol{\theta}_c(t))=\text{tr} \left(\mathbf{X} \mathbf{L}(\boldsymbol{\theta}_c(t))\mathbf{X}^\top\right)$ measure the similarity between rows and columns in image respectively. 

\noindent \textbf{Step 2}: \textit{Update parameters.} Minimize $\mathcal{L}_{a}$ by updating parameters $\boldsymbol{\theta}_a(t)$ with optimization algorithm such as Adam.

Iteration stops at $t=T$ when $\mathcal{L}_a(\boldsymbol{\theta}_a(T))$ is smaller than some given precision.

\noindent \textbf{Step 3}: \textit{Output  estimation.} The pixel value of $\mathbf{X}_{ij}$ on $\mathcal{G}\backslash \mathcal{X}$ is predicted by $\phi_{\boldsymbol{\theta}(T)}(\frac{i}{m},\frac{j}{n})$.

\noindent 
For high-dimensional data such as video, the regularizer captures the similarity between the vectorized frames.

\section{Additional experiments}
As a general image representation model, our method can be readily applied to other image tasks, including those higher dimensional ones. Table \ref{tab:denoising} shows that INRR outperforms INR in image denoising under variant noise types. Besides, we have also verified that INRR shows its power in video frame interpolation and RGB image inpainting. 
\begin{table}[!ht]
	\centering
	\caption{\footnotesize PSNR (dB) of denoised images by INR and INRR on (a) Baboon, (b) Man, (c) Barbara, (d) Boats and (e) Cameraman.}
	\label{tab:denoising}
	\vspace{-0.2cm}
	\tabcolsep=0.12cm
	\footnotesize
	\begin{tabular}{|l|l|l|l|l|l|l|} 
		\hline
		Noise Type     &Method  & (a) & (b)  & (c) & (d) & (e)  \\ 
		\hline
		\multirow{2}{*}{Gaussian ($\sigma=10$)}    & INR     & 28.5   & 29.3 & 30.5    & 30.8  & 33.1       \\ 
		\cline{2-7}
		& INRR    & \textbf{29.0}   & \textbf{29.7} & \textbf{30.6}    & \textbf{31.2}  & \textbf{33.3}       \\ 
		\hline
		\multirow{2}{*}{Salt \& Peper ($r=0.95$) } & INR     & 21.9   & 23.0 & 25.4    & 24.7  & 25.9       \\ 
		\cline{2-7}
		& INRR    & \textbf{22.0}   & \textbf{23.1} & \textbf{25.8}    & \textbf{24.9}  & \textbf{25.9}       \\ 
		\hline
		\multirow{2}{*}{Poisson ($\lambda=50$)}     & INR     & 23.2   & 24.9 & 27.0    & 25.9  & 27.7       \\ 
		\cline{2-7}
		& INRR    & \textbf{23.6}   & \textbf{25.2} & \textbf{27.1}    & \textbf{26.3}  & \textbf{28.1}       \\
		\hline
	\end{tabular}
	\vspace{-0.2cm}
\end{table}

Here we test INRR on video interpolation, where the data is represented by $\phi_{\boldsymbol{\theta}}(x,y,t)=[r,g,b]:\mathbb{R}^3\mapsto \mathbb{R}^3$.
Each frame is an RGB image, as shown in \Figref{fig:timeframe} (a). 
The tested video is a scene of water droplets that owns 202 frames.
We sample 21 frames uniformly as training data; the rest are test data.
\Figref{fig:timeframe} (b) shows that INRR can capture the non-local self-similarity between different frames. 
The average PSNR of INRR is 37.5 dB, while vanilla INR is 36.8 dB. Furthermore, we have validated the inpainting performance on a dataset BSD100 \cite{DavidMartin2001ADO} which includes 100 RGB images. 
The sampling mode is the same as in Figure 2(a) in the paper. 
The average PSNR of INRR is 28.8 dB, while vanilla INR is 27.1 dB. 

\begin{figure}[htb]
	\centering
	\includegraphics[width=\linewidth]{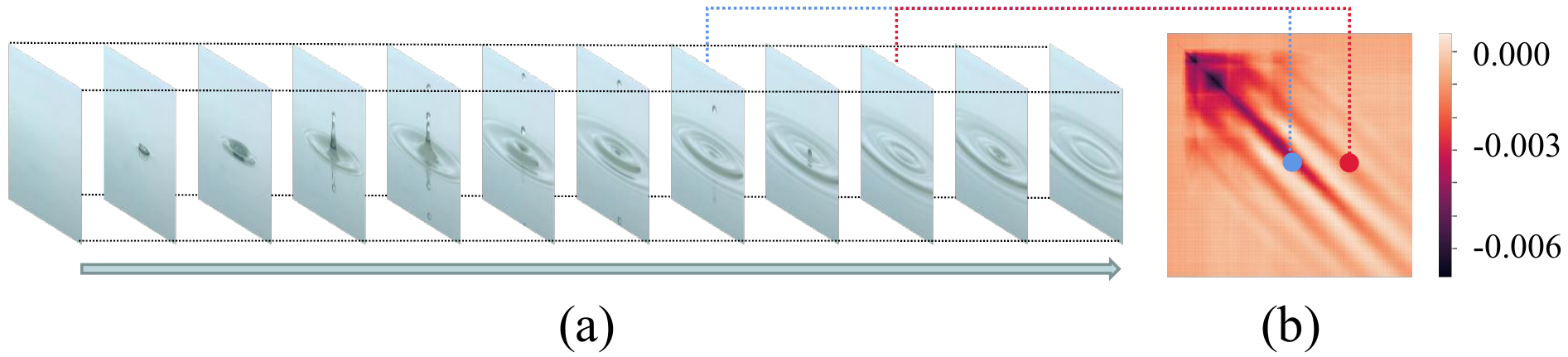}
	\vspace{-0.7cm}
	\caption{\footnotesize The similarity between different frames (a) can be captured by $\mathbf{L}$ (L392) in INRR (b) which is helpful to frame interpolation.}
	\vspace{-0.4cm}
	\label{fig:timeframe}
\end{figure}

\section{Smoothness of Laplacian matrix}
\begin{figure}[htb]
	\centering
	\includegraphics[width=\linewidth]{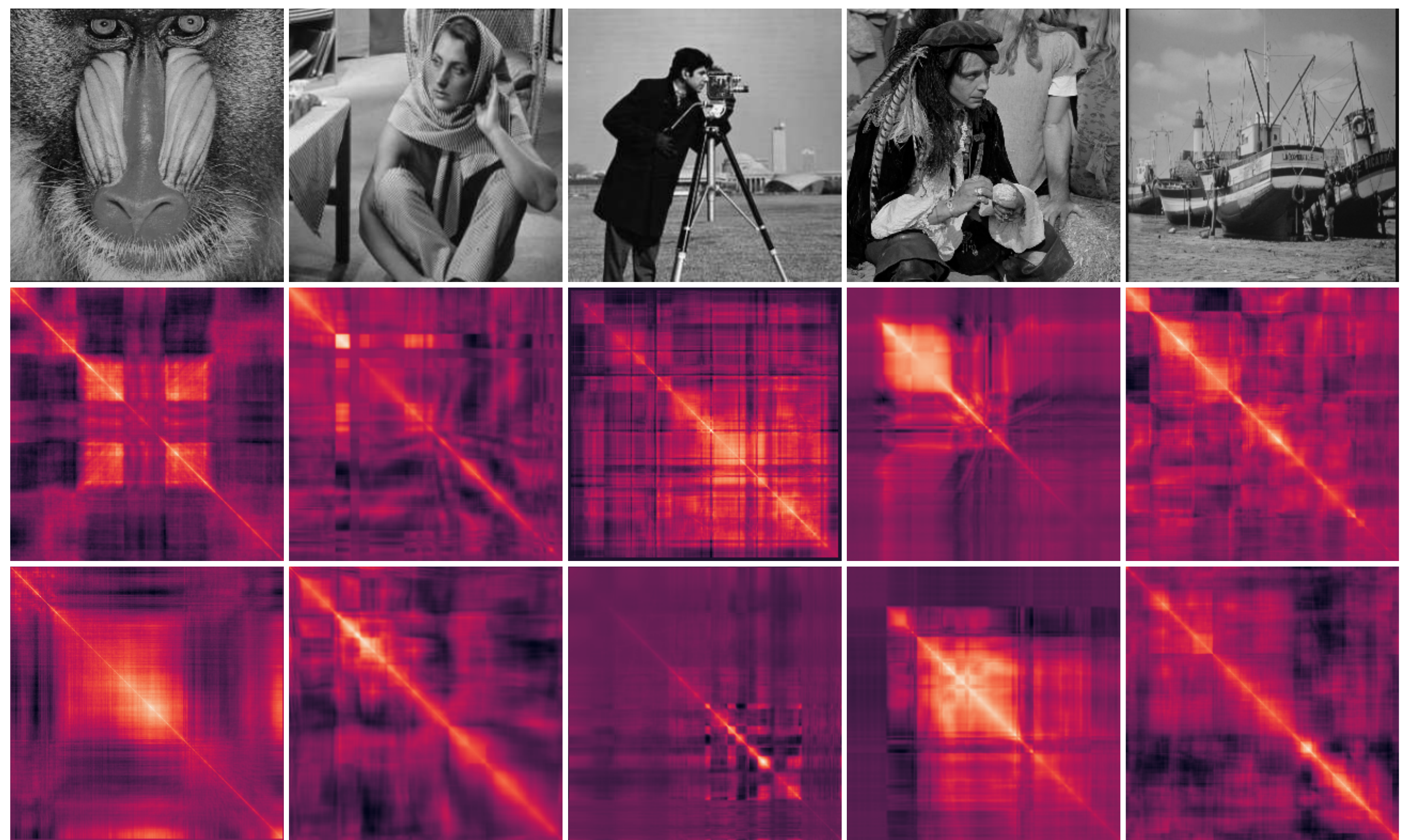}
	\caption{The first row show five different $256\times 256$ gray-scale images. The second row shows the column covariance matrix, and the third row shows the row covariance matrix.}
	\label{fig:app_smooth}
\end{figure}
We first calculate the covariance matrix of the columns and the rows in \Figref{fig:app_smooth}. The covariance matrix of $\mathbf{X}$ is $\mathbf{C}(\mathbf{X})$, where $\mathbf{C}_{ij}=\mathbb{E}[\mathbf{X}_{:,i}-\mathbb{E}(\mathbf{X}_{:,i})][\mathbf{X}_{:,j}-\mathbb{E}(\mathbf{X}_{:,j})]$, which measures the similarity among the columns. While the similarity among rows is $\mathbf{C}(\mathbf{X}^\top)$. As we can see, all the covariance matrices of various images are locally smoothly. Dong et al. proposed to utilize the smoothness of the Laplacian matrix by an extra regularizer \cite{BinDong2019CURECR}. We use an INR to encode the smoothness in such a Laplacian matrix implicitly; that is, our proposed INRR combines the self-similarity and smoothness of the Laplacian matrix at the same time.

Because $g(\boldsymbol{\theta};\mathbf{u})$ is an INR which is a smooth FCN about $\mathbf{u}$, $\mathbf{A}(\boldsymbol{\theta})=\frac{\text{exp}\left(g^\top(\boldsymbol{\theta};\mathbf{u})g(\boldsymbol{\theta};\mathbf{u})\right)}{\mathbf{1}^\top \text{exp}\left(g^\top(\boldsymbol{\theta};\mathbf{u})g(\boldsymbol{\theta};\mathbf{u})\right)\mathbf{1}}$ is smooth according to its expression. It means that a slight change of $\mathbf{u}$ generally leads to a slight change of $\mathbf{L}(\boldsymbol{\theta})=\mathbf{D}(\boldsymbol{\theta})-\mathbf{A}(\boldsymbol{\theta})$, which can be controlled by a Lipschitz constant. So we conclude that $\mathbf{A}(\boldsymbol{\theta})$ smoothes $\mathbf{L}(\boldsymbol{\theta})$, which is different from the vanilla $\mathbf{L}$. Furthermore, we can deduce a conclusion similar to Theorem 1 that the smoothness of $\mathbf{L}(\boldsymbol{\theta})$ is controlled by the $\delta$ of $g(\boldsymbol{\theta};\cdot)$. Smaller $\delta$ leads to a smoother result. Moreover, INRR degenerates to AIR when $\delta\rightarrow \infty$.

\section{Implicit bias}
\begin{figure}[htb]
	\centering
	\includegraphics[width=\linewidth]{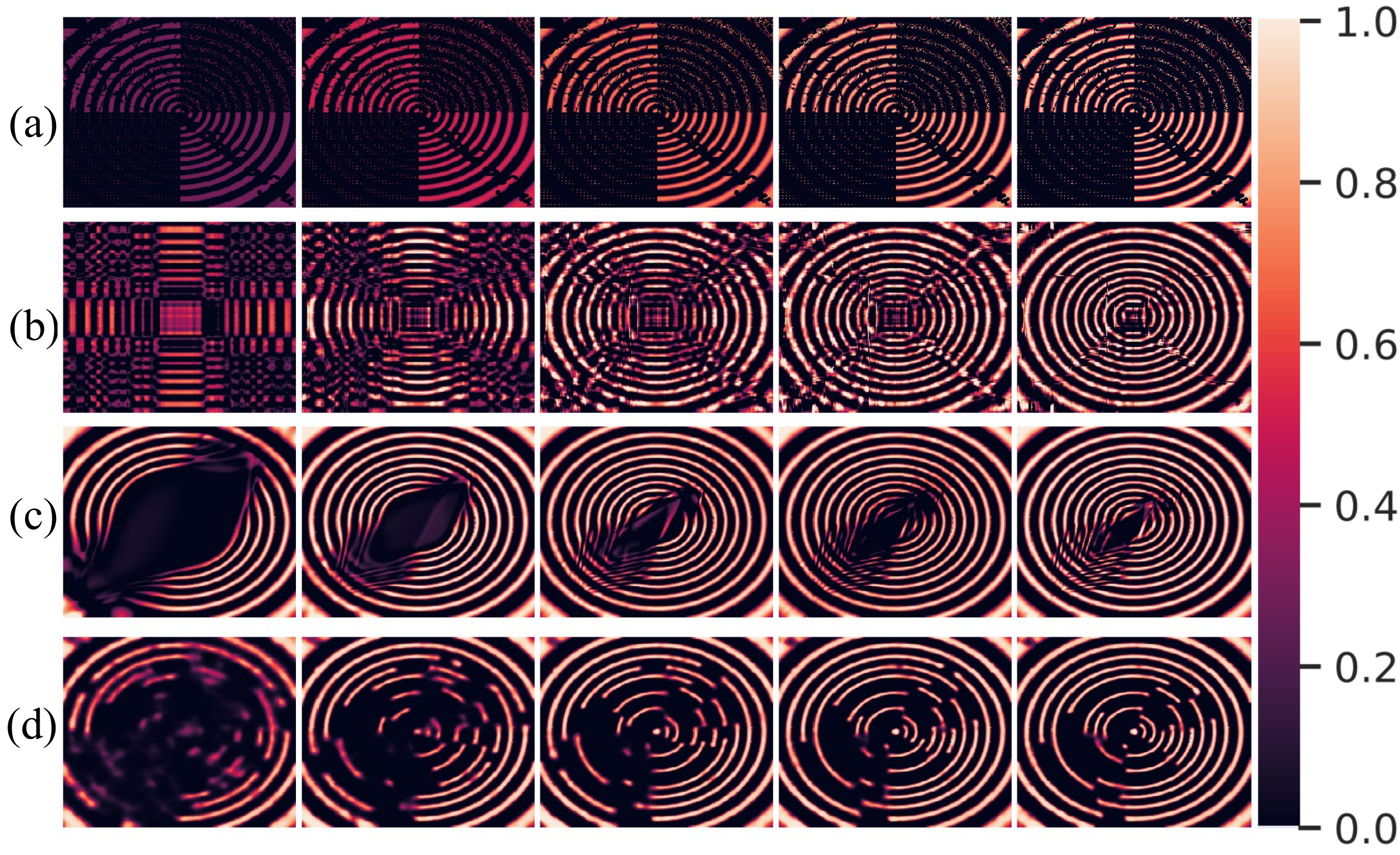}
	\caption{Fitting a $256\times 256$ synthetic data with a combine missing at different training step with (a) single matrix (DMF with only one factor), (b) DMF with three factors, (c) ReLU neural network, and (d) SIREN. The effective rank of the fitted matrix is shown in (e).}
	\label{fig:bias_comple}
\end{figure}
\begin{figure}[htb]
	\centering
	\includegraphics[width=\linewidth]{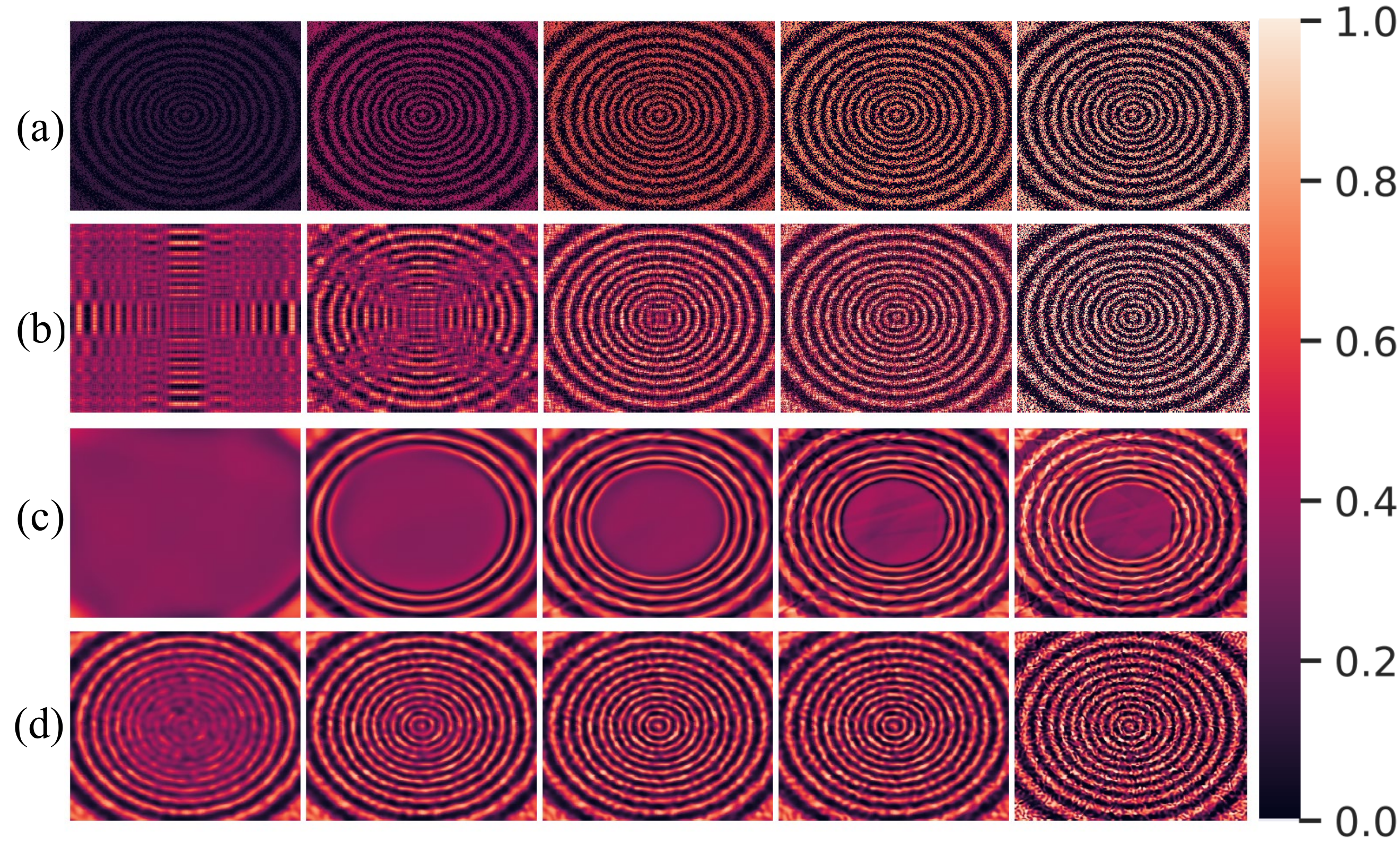}
	\caption{Fitting a $256\times 256$ synthetic data with Gaussian noise at different training step with (a) single matrix (DMF with only one factor), (b) DMF with three factors, (c) ReLU neural network, and (d) SIREN. The effective rank of the fitted matrix is shown in (e).}
	\label{fig:bias_noise}
\end{figure}
The neural network tends to converge to a good solution and may suffer from over-fitting with the training goes. Researchers explain this phenomenon by the implicit bias of neural networks. We show the implicit bias by fitting the synthetic data, which is sampled from function $s(x,y)=\sin\left(25\pi\sin\left(\frac{\pi}{3}\cdot \sqrt{x^2+y^2}\right)\right)$, where $\{(x_i,y_j)|i,j\}$ is a uniform $256\times 256$ grid on $[-1,1]\times [-1,1]$, where the local frequency of the synthetic data increases from boundary to center. We show two tasks on such synthetic data: fitting the incomplete data and the noisy data in \Figref{fig:bias_comple} and \Figref{fig:bias_noise}, respectively.  

As \Figref{fig:bias_comple}(a) shows, the single layer DMF fits all pixels without bias, and the pixels of the fitted image increase gradually. While \Figref{fig:bias_comple}(b) shows the low-rank bias of the three-layer DMF fitting the synthetic data from low-rank to high-rank. \Figref{fig:bias_comple}(c,d) shows that a fully connected neural network's bias is related to the data frequency and sampling rate. With the bias mentioned above, it is possible to complete an image without an extra explicit regularizer. Similarly, these neural network has similar phenomenon when fitting the noisy data.

\section{Recovered image}
\begin{figure}[htb]
	\centering
	\includegraphics[width=\linewidth]{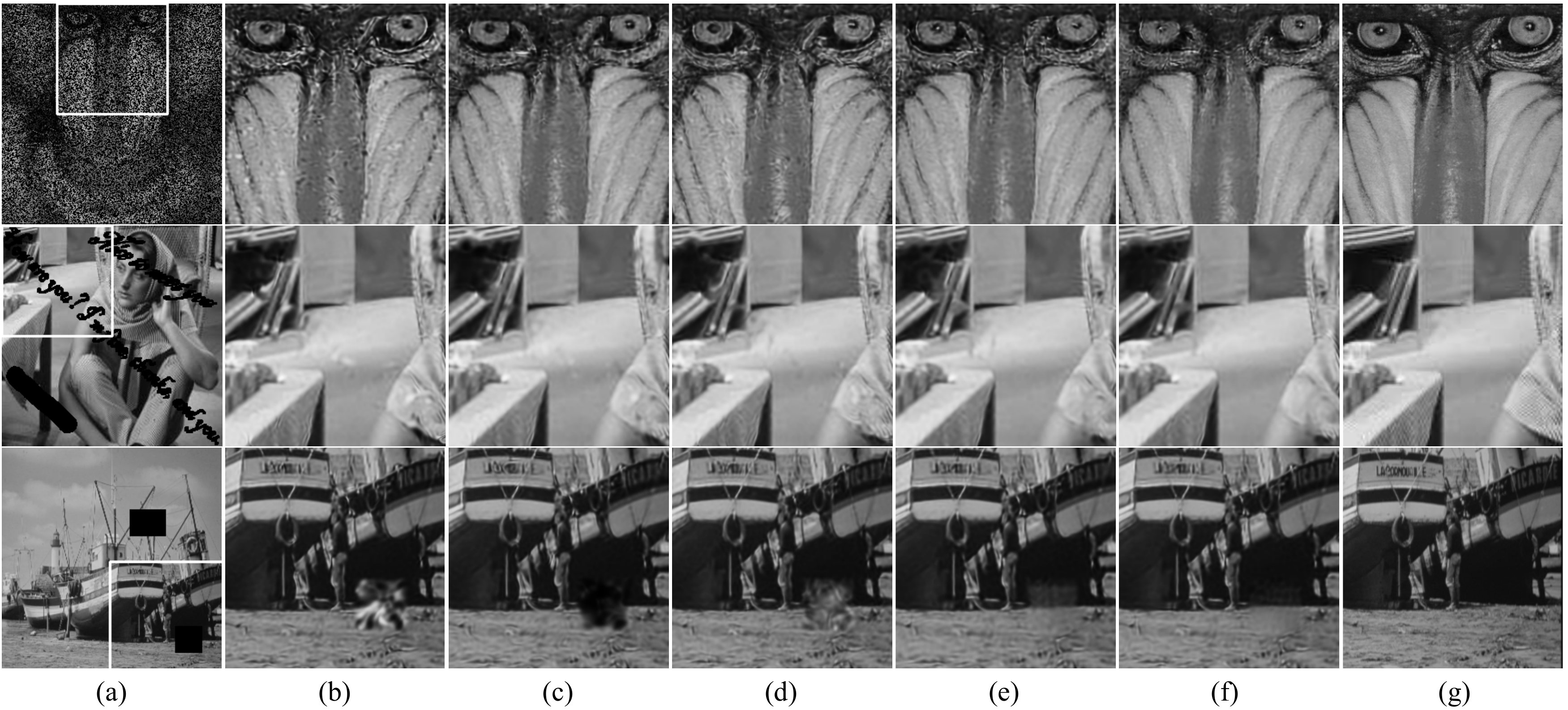}
	\caption{Result of image inpainting with three types of missing data by different regularized INR including (b) INR without regularization, (c) with TV, (d) $L_2$, (e) AIR, (f) INRR, and (g) original image. The hyper-parameters of benchmark models and algorithms are adopted from the original paper.}
	\label{fig:zoom_ori}
\end{figure}
We show the recovered image in \Figref{fig:zoom_ori}.

\fi

\end{document}

%% file: cvpr_header.tex
\usepackage[pagenumbers]{cvpr}
\usepackage{graphicx}
\usepackage{amsmath}
\usepackage{amssymb}
\usepackage{booktabs}

\input{_macros.tex}  

\usepackage{xr-hyper}

\makeatletter
\newcommand*{\addFileDependency}[1]{
  \typeout{(#1)}
  \@addtofilelist{#1}
  \IfFileExists{#1}{}{\typeout{No file #1.}}
}

\makeatother

\usepackage[pagebackref,breaklinks,colorlinks]{hyperref}
\usepackage[capitalize]{cleveref}
\crefname{section}{Sec.}{Secs.}
\crefname{table}{Table}{Tables}
\crefname{figure}{Fig.}{Figs.}

%% file: _macros.tex
\usepackage{times}
\usepackage{microtype}
\usepackage{epsfig}
\usepackage[table,xcdraw]{xcolor}
\usepackage{caption}
\usepackage{float}
\usepackage{placeins}
\usepackage{color, colortbl}
\usepackage{stfloats}
\usepackage{enumitem}
\usepackage{tabularx}
\usepackage{xstring}
\usepackage{multirow}
\usepackage{xspace}
\usepackage{url}
\usepackage{subcaption}
\usepackage{xcolor}
\usepackage[hang,flushmargin]{footmisc}

\ifcamera \usepackage[accsupp]{axessibility} \fi

% ...

\ifarxiv  \fi

\newcommand{\R}[1]{{%
    \textbf{%
        \ifstrequal{#1}{1}{\textcolor{red}{R#1}}{%
        \ifstrequal{#1}{2}{\textcolor{blue}{R#1}}{%
        \ifstrequal{#1}{3}{\textcolor{magenta}{R#1}}{%
        \ifstrequal{#1}{4}{\textcolor{teal}{R#1}}{%
                           \textcolor{cyan}{R#1}%
        }}}}%
    }%
}}

%% file: _math.tex
\usepackage{amsmath,amsfonts,bm}
\newtheorem{thm}{Theorem}
\newtheorem{cor}{Corollary}

\newtheorem{proof}{Proof}

\def\Figref#1{Figure~\ref{#1}}
\def\Tabref#1{Table~\ref{#1}}

\def\eqref#1{equation~\ref{#1}}

\def\1{\bm{1}}

\def\mB{{\bm{B}}}

\def\mI{{\bm{I}}}

\DeclareMathAlphabet{\mathsfit}{\encodingdefault}{\sfdefault}{m}{sl}
\SetMathAlphabet{\mathsfit}{bold}{\encodingdefault}{\sfdefault}{bx}{n}

